\documentclass{article}
\usepackage{iclr2027_conference,times}
\usepackage{amsmath,amssymb}
\usepackage{graphicx}
\usepackage{float}
\usepackage{wrapfig, needspace}
\usepackage{booktabs,array}
\usepackage{colortbl}
\definecolor{d2hoirow}{gray}{0.93}
\usepackage{pifont}
\usepackage[hidelinks]{hyperref}
\usepackage{url}

\iclrfinalcopy

\newcommand{\DATASET}{SituHOI}
\newcommand{\BENCHMARK}{FlexHOI-Bench}
\newcommand{\METHOD}{SharedHOI}
\newcommand*{\AUTHOREMAIL}{}

\title{\DATASET: Scalable Capture of Scene-Situated\\
Full-Body Human-Object Interaction}
\author{\textbf{Qifan Wang\textsuperscript{\dag}\hspace{1.3em}
Qihang Fang\textsuperscript{\dag}\hspace{1.3em}
Boao Shi\hspace{1.3em}
Yushi Du}\\
\textbf{Pei Zhou\hspace{1.3em}
Siyan Dong\hspace{1.3em}
Yanchao Yang\textsuperscript{*}}\\[10pt]
The University of Hong Kong\\[3pt]
\textsuperscript{\dag}Equal contribution\hspace{1.3em}\textsuperscript{*}Corresponding author
\ifx\AUTHOREMAIL\empty\else\\[3pt]{\ttfamily\AUTHOREMAIL}\fi}
\makeatletter
\renewcommand{\@maketitle}{
\vbox{
\hsize\textwidth
\centering
\setlength{\parskip}{0pt}
\hrule height 4pt
\vspace{16pt}
{\normalfont\bfseries\fontsize{17.28}{20}\selectfont\@title\par}
\vspace{14pt}
\hrule height 1pt
\vspace{23pt}
{\normalfont\fontsize{10}{12}\selectfont\@author\par}
\vspace{18pt}
}
}
\makeatother
\hypersetup{
pdftitle={SituHOI: Scalable Capture of Scene-Situated Full-Body Human-Object Interaction},
pdfauthor={Qifan Wang, Qihang Fang, Boao Shi, Yushi Du, Pei Zhou, Siyan Dong, Yanchao Yang},
pdfsubject={Scene-grounded human-object interaction dataset and motion prediction benchmark},
pdfkeywords={SituHOI, human-object interaction, RGB-D, motion prediction}
}

\begin{document}
\maketitle
\fancyhead{}
\renewcommand{\headrulewidth}{0pt}

\begin{abstract}
Modeling human-object interaction (HOI) in real environments requires jointly representing whole-body motion, manipulated objects, and scene geometry, yet existing datasets trade off scalable capture against structured, scene-aligned supervision. We introduce \DATASET{}, 
a large-scale RGB-D dataset containing 18,429 annotated clips across 16 indoor rooms and approximately 86 hours of quality-screened activity. 
It pairs recordings from a single moving RGB-D camera with static scene scans, object geometry, language descriptions, and estimated human and object trajectories. Using the scans as persistent references, our pipeline combines camera localization and depth-based metric alignment with human mesh recovery, object tracking, and temporal refinement to register all geometric annotations in a shared floor-aligned frame. 
These data support \BENCHMARK{}, a goal-conditioned joint motion prediction benchmark with variable observation and prediction durations. 
Given observed histories, scene and object geometry, and a terminal object pose, models predict future human and object motion. 
We provide \METHOD{} as a reference predictor and evaluate world-space accuracy and human-object coordination. 
On a MoCap subset, our pipeline reduces human G-MPJPE from 195.4 to 68.3 mm and object translation error from 43.0 to 28.7 mm. SharedHOI leads four adapted baselines on all three aggregate FlexHOI-Bench metrics.
Data will be released upon acceptance. An anonymized code repository is available at https://anonymous.4open.science/r/iclr2027-anonymous-code-1DFB.
\end{abstract}

\begingroup
\setlength{\parskip}{4pt}
\section{Introduction}
\label{sec:introduction}

Recent advances in human motion generation have enabled increasingly realistic and controllable whole-body motion~\citep{tevet2023mdm,karunratanakul2023gmd}. 
However, the growing interest in humanoid robots and embodied agents is driving a shift from modeling human motion in isolation to scene-grounded human-object interaction~\citep{he2025omnih2o,xu2025intermimic}. 
In this setting, human motion must respect scene geometry and contact constraints while remaining temporally coordinated with the motion of manipulated objects~\citep{diller2024cghoi,li2024chois}, 
making the problem substantially more challenging than conventional motion generation.

Learning such interactions requires data spanning diverse scenes, objects, and motions~\citep{bhatnagar2022behave,kim2025parahome}. 
Egocentric datasets collected with head-mounted cameras provide a convenient means of recording hand-object interactions~\citep{liu2022hoi4d,banerjee2025hot3d,hoque2026egodex}, 
but their restricted field of view and frequent occlusions make accurate full-body motion recovery difficult without additional sensing. Other datasets rely on marker-based motion-capture (MoCap) systems, 
which provide accurate motion measurements but require costly, instrumented environments~\citep{taheri2020grab}, 
or calibrated third-person multiview RGB systems, which demand substantial camera setup and provide no direct depth measurements~\citep{zhang2023neuraldome}. 
Existing capture paradigms therefore trade off scalability and scene diversity against full-body coverage and geometric accuracy.

RGB-D cameras offer a promising balance between flexible visual capture and direct metric depth.  
Existing full-body RGB-D interaction datasets typically employ multiple fixed cameras to improve coverage and reduce occlusion~\citep{bhatnagar2022behave,huang2024intercap,kogashi2026mmhoi}, 
but these systems require installation, synchronization, and calibration in each new environment 
, limiting long-form collection across diverse indoor spaces. 
A single moving RGB-D camera provides a more deployable alternative by following interactions beyond a fixed volume while retaining metric depth. 
However, its moving viewpoint and lack of multiview constraints make camera, human, and object motion difficult to disentangle and register consistently in a common scene frame.

To address this challenge, we develop a capture and reconstruction pipeline that separates static-scene reconstruction from dynamic-activity recording. 
Each room is scanned once to construct a metric, gravity-aligned scene model, after which activities are recorded with a single handheld RGB-D camera. 
We combine scene-referenced camera localization with depth-based human fitting, object-tracking recovery, and temporal refinement to obtain aligned motion estimates from the moving viewpoint. 
The protocol replaces per-scene multiview setup with a one-time scene scan and object-asset preparation, and requires no body- or object-mounted trackers.

Using this pipeline, we introduce \textbf{\emph{\DATASET}}, a large-scale RGB-D dataset of scene-grounded full-body human-object interaction. 
Curated from roughly 90 hours of raw recordings, \DATASET~comprises approximately 86 hours of activity across 16 indoor rooms, segmented into 18,429 language-annotated clips, including 10,335 dynamic-object clips covering 48.64 hours. 
Each recording includes RGB-D observations, reconstructed scene geometry, camera poses, and object assets, while each clip includes a natural-language description, estimated full-body human motion, and, where applicable, a rigid-object trajectory. All geometric annotations share a metric scene frame. To our knowledge, \DATASET~is the first full-body HOI dataset to provide this combination at scale using a single moving RGB-D camera without body- or object-mounted trackers (Tab.~\ref{tab:dataset_comparison}).

Building on \DATASET{}, we introduce \textbf{\BENCHMARK{}} for
\emph{goal-conditioned joint human-object motion prediction}
across variable observation and forecast durations.
Given motion histories, aligned scene and object geometry, and a terminal
object pose, a single model predicts future human and object motion
for the requested time horizons.
We further propose \textbf{\METHOD}, a unified predictor of both motions and their relative coordination, and evaluate it using complementary world-space and body-relative metrics that capture global accuracy and interaction coherence.

In summary, our contributions are:
\begingroup
\setlength{\topsep}{2pt}
\setlength{\itemsep}{2pt}
\setlength{\parsep}{0pt}
\setlength{\partopsep}{0pt}
\begin{itemize}

\item \emph{A large-scale, scene-grounded full-body HOI dataset.}
We introduce \DATASET, comprising approximately 86 hours of RGB-D activity across 16 indoor rooms and 18,429 language-annotated intervals, including 10,335 intervals involving dynamic objects. The dataset provides reconstructed room-scale scenes and estimated camera, human, and rigid-object trajectories in a shared metric scene frame.

\item \emph{A scalable single-camera capture and reconstruction pipeline.}
By separating one-time scene reconstruction from activity recording, our pipeline recovers temporally consistent camera, full-body human, and rigid-object motion from a single moving RGB-D stream. It requires neither a calibrated multi-camera rig nor body- or object-mounted trackers, facilitating deployment across diverse indoor environments.

\item \emph{A unified benchmark and model for goal-conditioned joint motion prediction.}
We introduce \BENCHMARK, which maps observed human-object motion, scene and object geometry, and a target object pose to the future motions of both the human and the object. We further propose \textbf{\METHOD} and evaluate it using complementary world-space and body-relative metrics that assess both global motion accuracy and human-object coordination.

\end{itemize}
\endgroup

\begin{table}[!htb]
\centering
\definecolor{datasetcheckgreen}{rgb}{0.10,0.45,0.25}
\definecolor{datasetcrossred}{rgb}{0.75,0.16,0.16}
\newcommand{\datasetcheck}{\textcolor{datasetcheckgreen}{\ding{51}}}
\newcommand{\datasetcross}{\textcolor{datasetcrossred}{\ding{55}}}
\caption{Comparison of capture setups, hours, sequence/segment counts,
environments (Envs.), and annotations.
Full-body 3D includes arms and legs; Object pose denotes dynamic-object
position and orientation. Scene geom. includes point clouds and layouts; Scene mesh requires environment or aligned static-furniture meshes,
not isolated manipulated-object meshes. Semantics covers action labels and
language descriptions. Uninstr. excludes body/object tracking attachments,
but permits capture cameras and calibration targets.
\datasetcheck: documented support; \datasetcross: criterion unmet or support
unestablished; NR: not reported. Definitions and sources:
Appendix~\ref{app:comparison}.}
\label{tab:dataset_comparison}
\begingroup
\scriptsize
\setlength{\tabcolsep}{0pt}
\renewcommand{\arraystretch}{1}
\newcolumntype{D}[1]{>{\raggedright\rightskip=2pt plus1fil\arraybackslash}m{#1}}
\newcolumntype{C}[1]{>{\centering\arraybackslash}m{#1}}
\begin{tabular}{@{}D{0.130\linewidth}D{0.150\linewidth}D{0.103\linewidth}D{0.107\linewidth}C{0.050\linewidth}C{0.085\linewidth}C{0.070\linewidth}C{0.071\linewidth}C{0.070\linewidth}C{0.090\linewidth}C{0.074\linewidth}@{}}
\toprule
\textbf{Dataset} & \textbf{Capture} & \textbf{Hours} & \textbf{Seq./seg.} & \textbf{Envs.} & \textbf{Full-body}\newline\textbf{3D} & \textbf{Object}\newline\textbf{pose} & \textbf{Scene}\newline\textbf{geom.} & \textbf{Scene}\newline\textbf{mesh} & \textbf{Semantics} & \textbf{Uninstr.} \\
\midrule
\multicolumn{11}{@{}l}{\textit{Full-body human-scene capture}} \\
PROX & 1 RGB-D & NR & 60 seq. & 12 & \datasetcheck & \datasetcross & \datasetcheck & \datasetcheck & \datasetcross & \datasetcheck \\
RICH & 6--8 RGB & NR & 142 seq. & 5 & \datasetcheck & \datasetcross & \datasetcheck & \datasetcheck & \datasetcheck & \datasetcheck \\
\midrule
\multicolumn{11}{@{}l}{\textit{Full-body interaction capture}} \\
GRAB & 54-cam MoCap & NR & 1,334 seq. & NR & \datasetcheck & \datasetcheck & \datasetcross & \datasetcross & \datasetcheck & \datasetcross \\
BEHAVE & 4 RGB-D & NR & 321 seq. & 5 & \datasetcheck & \datasetcheck & \datasetcross & \datasetcross & \datasetcross & \datasetcheck \\
InterCap & 6 RGB-D & NR & 223 seq. & 1 & \datasetcheck & \datasetcheck & \datasetcross & \datasetcross & \datasetcross & \datasetcheck \\
CHAIRS & 4 RGB-D +\newline hybrid MoCap & 17.3 & 1,390 seq. & NR & \datasetcheck & \datasetcheck & \datasetcross & \datasetcross & \datasetcheck & \datasetcross \\
OMOMO & 12-cam MoCap & $\sim$10 & NR & NR & \datasetcheck & \datasetcheck & \datasetcross & \datasetcross & \datasetcheck & \datasetcross \\
IMHD\textsuperscript{2} & 32 RGB + IMUs & NR & 295 seq. & NR & \datasetcheck & \datasetcheck & \datasetcross & \datasetcross & \datasetcross & \datasetcross \\
HOI-M\textsuperscript{3} & 42 RGB + IMUs & $>$20 & 199 seq. & 5 & \datasetcheck & \datasetcheck & \datasetcheck & \datasetcheck & \datasetcross & \datasetcross \\
HIMO & 20-cam MoCap\newline + gloves & 9.44 & 3,376 seq. & NR & \datasetcheck & \datasetcheck & \datasetcross & \datasetcross & \datasetcheck & \datasetcross \\
ParaHome & 70 RGB +\newline suit/gloves & 8.10 & 207 seq. & 1 & \datasetcheck & \datasetcheck & \datasetcheck & \datasetcheck & \datasetcheck & \datasetcross \\
HUMOTO & 2 RGB-D +\newline suit/gloves & 2.19 & 735 seq. & NR & \datasetcheck & \datasetcheck & \datasetcheck & \datasetcheck & \datasetcheck & \datasetcross \\
MMHOI & 4 RGB-D & NR & NR & NR & \datasetcheck & \datasetcheck & \datasetcross & \datasetcross & \datasetcheck & \datasetcheck \\
ACE-Data-0 & 8 exo + 4 ego\newline + MoCap & 150 & 75,000 ep. & 2 & \datasetcheck & \datasetcheck & \datasetcheck & \datasetcross & \datasetcheck & \datasetcross \\
HiPHI & Optical MoCap & 245.7 & NR & NR & \datasetcheck & \datasetcheck & \datasetcross & \datasetcross & \datasetcheck & \datasetcross \\
\midrule
\multicolumn{11}{@{}l}{\textit{Consolidated / augmented benchmark}} \\
InterAct / -X & 7 datasets +\newline optimization & 21.81 base\newline 30.70 aug. & 11,350 base\newline 16,201 aug. & NR & \datasetcheck & \datasetcheck & \datasetcross & \datasetcross & \datasetcheck & \datasetcross \\
\midrule
\multicolumn{11}{@{}l}{\textit{Egocentric hand-object interaction}} \\
HOI4D & Egocentric\newline RGB-D & NR & 4,000 seq. & 610 & \datasetcross & \datasetcheck & \datasetcheck & \datasetcross & \datasetcheck & \datasetcheck \\
HOT3D & Aria / Quest 3\newline + MoCap & 13.88 & NR & 1 & \datasetcross & \datasetcheck & \datasetcheck & \datasetcross & \datasetcross & \datasetcross \\
EgoDex & Vision Pro & 829 & 338,000 ep. & NR & \datasetcross & \datasetcross & \datasetcross & \datasetcross & \datasetcheck & \datasetcheck \\
\midrule
\rowcolor{d2hoirow}
\textbf{\DATASET{}} & \textbf{1 RGB-D}\newline\textbf{+ prior scans} & \textbf{$\sim$86 total}\newline\textbf{48.64 dyn.} & \textbf{10,335 seg.}\newline\textbf{(dynamic)} & \textbf{16} & \datasetcheck & \datasetcheck & \datasetcheck & \datasetcheck & \datasetcheck & \datasetcheck \\
\bottomrule
\end{tabular}
\par
\endgroup
\vspace{-12pt}
\end{table}

\vspace{-1ex}
\endgroup

\begingroup
\setlength{\parskip}{4pt}
\vspace{-0.5ex}
\section{Related Work}
\label{sec:related_work}

\vspace{-0.5ex}
\paragraph{Scene-grounded human-object interaction datasets.}
Existing datasets cover scene-aware full-body motion and egocentric body capture in reconstructed environments~\citep{hassan2019prox,huang2022rich,zhang2022egobody}. Other resources capture full-body human-object interaction using motion capture or multiview RGB-D~\citep{taheri2020grab,lu2025humoto,bhatnagar2022behave,huang2022intercap}, or focus on articulated and egocentric hand-object manipulation~\citep{fan2023arctic,liu2022hoi4d,banerjee2025hot3d,hoque2026egodex}. Recent datasets further extend interaction capture to contextual environments, collaborative activities, and multiple humans or objects~\citep{zhang2024hoim3,kim2025parahome,liu2025core4d,kogashi2026mmhoi}. Across these efforts, complete human and object motion, explicit scene geometry, semantic supervision, and portable capture remain difficult to obtain jointly. \DATASET{} combines these properties through single-camera RGB-D observations, scene-aligned motion, room geometry, and language.

\vspace{-1.5ex}
\paragraph{Scene-aligned interaction reconstruction.}
Constructing scene-aligned interaction data requires camera and scene reconstruction~\citep{schonberger2016sfm,sarlin2019hloc,fang2022towards}, expressive body recovery~\citep{pavlakos2019smplx,lin2023osx}, and object segmentation and tracking~\citep{ravi2025sam2,wen2024foundationpose}. These components provide complementary estimates but do not automatically share consistent scales, coordinate frames, or temporal conventions. Our pipeline integrates them through depth-based alignment, floor referencing, temporal refinement, and quality filtering to produce a unified metric representation.

\vspace{-1.5ex}
\paragraph{Human-motion  interaction modeling.}
Human-object models forecast interactions from histories, generate human motion from prescribed object trajectories, or jointly synthesize human and object motion conditioned on language, contacts, or sparse waypoints~\citep{xu2023interdiff,li2023omomo,diller2024cghoi,li2024chois}. These formulations provide different amounts of future object information. \BENCHMARK{} instead conditions on  interaction history, aligned scene and object geometry, and only a terminal object pose, while requiring both future human and object motion. Its world-space and body-relative evaluation measures global accuracy and interaction coordination.
\par
\endgroup

\begingroup
\setlength{\parskip}{4pt}
\section{\DATASET{} Dataset}
\label{sec:dataset}
\label{sec:dataset_design}

\begin{figure}[!t]
\centering
\includegraphics[width=\linewidth
,pagebox=cropbox]{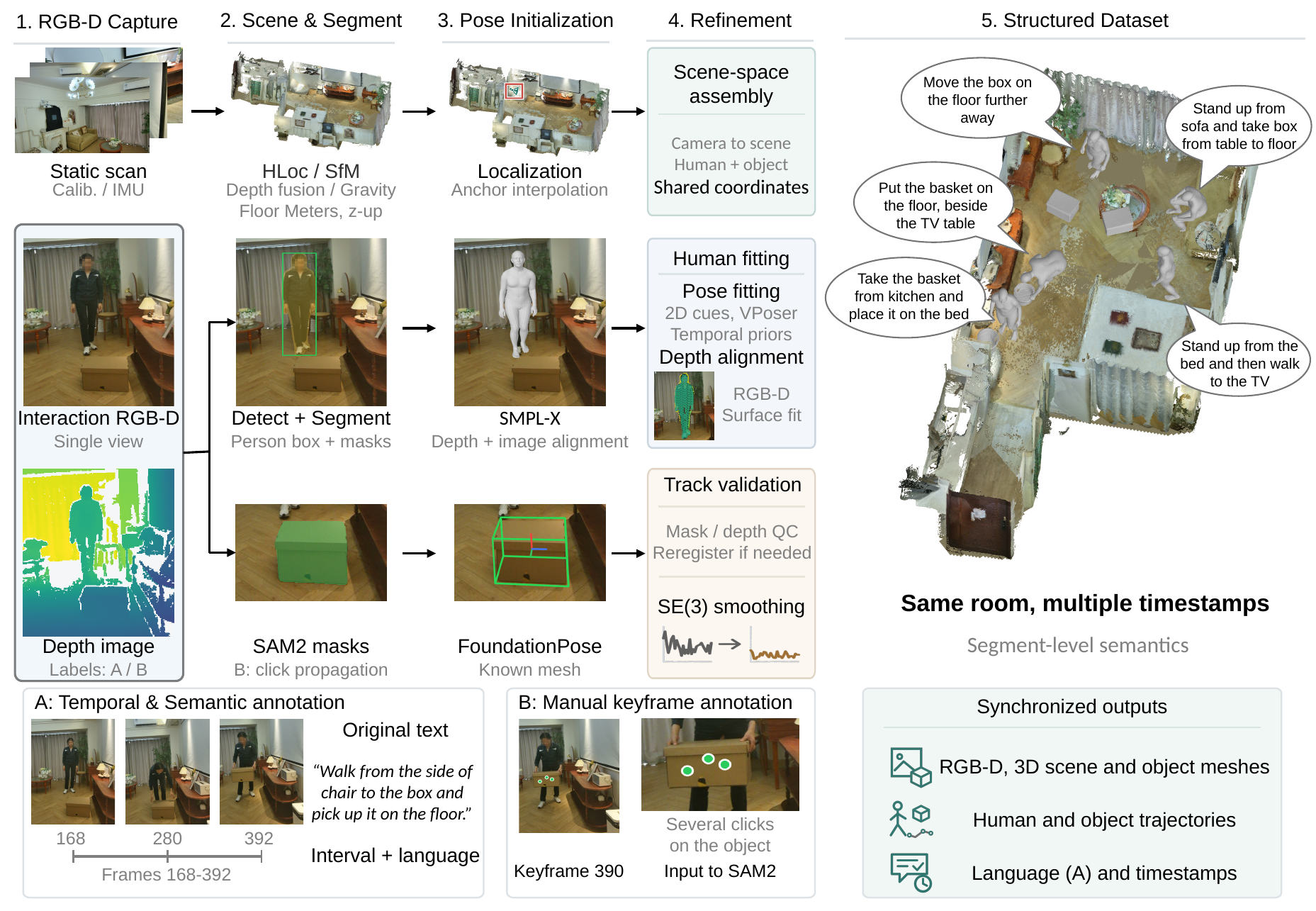}
\caption{\DATASET{} construction pipeline. We collect room scans, RGB-D activity, object meshes, and manual annotations, which support camera, human, and object estimation in a shared scene space. Annotations specify activity intervals, language descriptions, and target-object clicks. 
}
\label{fig:dataset_pipeline}
\vspace{-16pt}
\end{figure}

To capture scene-grounded human-object interactions with a single handheld
RGB-D camera, we collect three complementary types of data: 
static room scans
for scene reconstruction and camera localization, 
dedicated object scans for object mesh reconstruction and pose tracking, 
and interaction recordings for human and object tracking (Fig.~\ref{fig:dataset_pipeline}). 
With this pipeline, we build a novel dataset \textbf{\DATASET{}} with RGB-D observations,
human and object trajectories, and geometry in a shared metric frame for each
of 16 indoor rooms. 
After screening, the dataset contains 18,429 language-annotated clips spanning
approximately 86 hours, of which 10,335 clips covering 48.64 hours involve
13 dynamic objects, and the remaining clips capture interactions with static
furnitures.

\subsection{Static room scans and scene reconstruction}
\label{sec:dataset_scene}

For each room, we capture a static RGB-D scan and retain the sensor
calibration and available IMU measurements. 
We use
HLoc~\citep{sarlin2019hloc} for image retrieval and feature matching
across the scan, 
and COLMAP~\citep{schonberger2016sfm} to recover the
scan-camera poses and sparse scene structure from these correspondences. Since COLMAP cannot recover the correct metric scale, 
we estimate the metric scale from the median of the ratio between captured depth and COLMAP-recovered depth, then fuse the backprojected observations into a dense mesh. 
We estimate the vertical direction from camera orientations and reliable IMU data,
fit the floor plane to 3D scene points, and align it
with $z=0$, with the vertical direction pointing along $+z$.
The resulting metric, gravity-aligned reconstruction provides both
room geometry and a reference map for localizing the interaction camera.
Further details are given in Appendix~\ref{app:scene_details}.

\subsection{Dedicated object scans and mesh reconstruction}
\label{sec:dataset_object_mesh}

For each dynamic object, we acquire a dedicated RGB-D scan and reconstruct
a textured mesh using BundleSDF~\citep{wen2023bundlesdf}. This scan is
collected separately from the interaction recordings so that the object's
geometry can be reconstructed before estimating its motion. The resulting
mesh serves as known geometry for rigid-pose registration and tracking in
the interaction recordings.

\subsection{Interaction recordings and trajectory reconstruction}
\label{sec:dataset_acquisition}
\label{sec:dataset_reconstruction}

\paragraph{Recording and annotation.}
We record dynamic activities across 16 rooms using a single handheld RGB-D
camera at 30 FPS, with RGB resolution of $1280\times960$ and depth resolution
of $640\times480$. We collect approximately 90 hours of raw recordings.
After excluding segments with severe person-object occlusion, motion blur,
or missing or corrupted depth, and omitting gaps between annotated
activities, the retained clips cover approximately 86 hours. This duration
measures annotated clip coverage rather than continuous raw recording time.

Annotators mark clip boundaries, write free-form English descriptions of
actions and spatial relations, and identify the involved objects and their
types. For each dynamic object interaction clip, 
they choose a keyframe
and provide one or more point clicks to initialize mask propagation. 
Fig.~\ref{fig:dataset_pipeline} shows an example in which a language
description covers frames 168-392 and three clicks on the box at frame 390
initialize its mask.

\paragraph{Camera localization.}
We localize the interaction camera against the metric scene map constructed
from the static room scan. By default, for every 30th recording frame, HLoc
retrieves relevant scan images, matches visual features, and associates
2D image pixels with 3D map points. Its localization pipeline then
estimates the camera pose using RANSAC-based PnP followed by refinement. 
The successfully localized frames serve as camera-pose anchors.
Camera poses between retained anchors are obtained by linear interpolation of translation and spherical linear interpolation (SLERP) of rotation; poses outside the anchor span use the nearest retained endpoint.
The resulting camera
trajectory places the human and object motion reconstructed below in the
same coordinate system.

\paragraph{Human motion.}
OSX~\citep{lin2023osx} provides initial SMPL-X pose and shape from RGB.
We get root translation by back-projecting person-region depth and
transform the body to the scene space using the camera pose, then apply
two-stage fitting.
The first stage
refines this estimate using confidence-weighted 2D keypoints, temporal
regularization, and the VPoser body-pose prior~\citep{pavlakos2019smplx}.
The second stage further aligns the body surface with 3D points back-projected from depth pixels within SAM~2 person masks~\citep{ravi2025sam2}.

Let $J_t$ and $V_t$ denote the SMPL-X joints and vertices at frame $t$, and
let $R_t$, $q_t$, and $z_t$ denote the optimized joint rotations, root
translation, and VPoser pose code. The two stages minimize
\begin{equation}
E_h^{(1)}
= \lambda_{2D}E_{2D}
+ E_{\mathrm{prior}}
+ E_{\mathrm{preserve}}
+ \sum_{x\in\{J,V,R,q,z\}}\lambda_x E_{\Delta^2 x_t},
\qquad
E_h^{(2)} = E_h^{(1)}+\lambda_D E_D.
\label{eq:human_fitting}
\end{equation}
Here, $\Delta^2x_t=x_{t+1}-2x_t+x_{t-1}$ is the second temporal difference.
$E_{2D}$ is a robust reprojection loss between projected SMPL-X joints and
confidence-weighted RTMW detections~\citep{jiang2024rtmw}, using 63 matched
body, foot, and hand landmarks. $E_D$ is a robust bidirectional Chamfer
distance between masked RGB-D person points and sampled body vertices.
$E_{\mathrm{prior}}$ regularizes body and hand pose, while
$E_{\mathrm{preserve}}$ discourages deviations from the OSX initialization.
For rotational variables, temporal differences are computed on
rotation-matrix entries. Appendix~\ref{app:human_details} gives the complete
loss definitions and optimization settings.

\paragraph{Object motion.}
\label{par:dataset_object_tracking}
We propagate object masks from the annotated keyframe clicks using
SAM~2~\citep{ravi2025sam2} and estimate the pose of each object using its
dedicated-scan mesh and FoundationPose~\citep{wen2024foundationpose}.
Our tracking safeguards include registering candidate poses on visible
frames around the annotated keyframe and selecting an initialization based
on mask overlap, depth agreement, and whether the object is truncated by
the image boundary. FoundationPose then tracks the object forward and
backward from the selected frame.

During tracking, we run a safeguard process to improve tracking robustness. Low overlap between the projected mesh and propagated mask
triggers a new registration proposal. We accept the proposal only if it
improves mask agreement and passes the depth, visibility, and
trajectory-continuity checks in Appendix~\ref{app:tracking_details}. The same
recovery procedure applies when an object reappears after an occlusion.
Rejected poses within visible spans are interpolated when bounded by
reliable estimates. During invisible spans, we retain the last pose as a
placeholder and mark those frames as unobserved.

We apply object refinement to visible trajectories to reduce temporal jitter
while keeping them close to the tracked poses. Let $p_t^0$ and $Q_t^0$ denote
the tracked scene-frame translation and rotation. We optimize bounded
corrections $\delta p_t$ and $\delta r_t$ such that
\begin{equation}
p_t=p_t^0+\delta p_t,
\qquad
Q_t=\operatorname{Exp}(\delta r_t)Q_t^0,
\end{equation}
where $\operatorname{Exp}$ maps an axis-angle correction to a rotation
matrix. Defining
\[
\omega_t=\operatorname{Log}(Q_{t+1}Q_t^\top),
\qquad
\Delta\omega_t=\omega_{t+1}-\omega_t,
\]
we minimize
\begin{equation}
E_o=
\lambda_p\operatorname{MSE}(\Delta^2p)
+\lambda_\omega\operatorname{MSE}(\Delta\omega)
+\lambda_{p0}\operatorname{MSE}(\delta p/s_p)
+\lambda_{r0}\operatorname{MSE}(\delta r/s_r).
\label{eq:object_refinement}
\end{equation}
The first two terms reduce translational and rotational jitter; the last
two penalize deviations from the tracked trajectory. We bound corrections
by $s_p=0.05$ m and $s_r=0.25$ rad, which also normalize the fidelity terms.
Corrections are fixed to zero on invisible frames, leaving their held
placeholder poses unchanged. Appendix~\ref{app:object_details} provides
the remaining optimization details.

\setlength{\parskip}{3pt}
\subsection{Quality control}
\label{sec:dataset_quality_control}

We first screen the recordings for severe occlusion, motion blur, and
missing or corrupted depth, as described above. After reconstruction, we
combine manual inspection with automated checks of source-data integrity,
camera-pose coverage, trajectory continuity, and the agreement of human
and object geometry with RGB-D masks and depth. Appendix~
\ref{app:quality_control} gives the checks and acceptance thresholds.
We further assess trajectory quality through component ablations and
independent synchronized MoCap validation in
Section~\ref{sec:mocap_validation}.

\vspace{-1ex}

\vspace{-1ex}
\subsection{Corpus coverage and comparison}
\label{sec:dataset_statistics}

\paragraph{Comparison with existing resources.}
Tab.~\ref{tab:dataset_comparison} highlights \DATASET{}'s combination of
single-camera RGB-D capture, full-body and dynamic-object trajectories,
room meshes, and language without body- or object-mounted trackers.
Their common scene frame links explicit surfaces and motion for geometric
analysis and scene-conditioned modeling (criteria and sources:
Appendix~\ref{app:comparison}).

\vspace{-1ex}
\paragraph{Object and clip distributions.}
Fig.~\ref{fig:dataset_statistics} shows object frequencies and clip durations.
The 13 types span furniture, containers, and smaller household objects;
boxes and chairs dominate across varied appearances and manipulations.
Longer clips support history/forecast resampling beyond current benchmark
windows; Appendix~\ref{app:corpus_audit} details the inventory, durations,
and language statistics.
\label{par:dataset_distributions_end}

\par
\begingroup
\setlength{\intextsep}{4pt}
\begin{figure}[!t]
\centering
\includegraphics[width=\linewidth, height=0.28\linewidth]{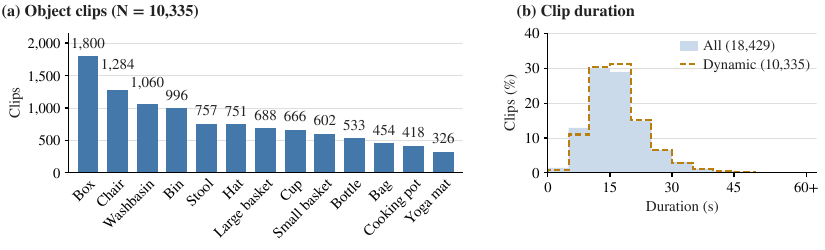}
\caption{
Dataset distributions: (a) object counts; (b) clip durations, normalized per group.
}
\label{fig:dataset_statistics}
\end{figure}
\endgroup
\endgroup
\allowbreak
\suppressfloats[t]
\begin{figure}[t]
\centering
\includegraphics[width=\linewidth, height=0.35\linewidth]{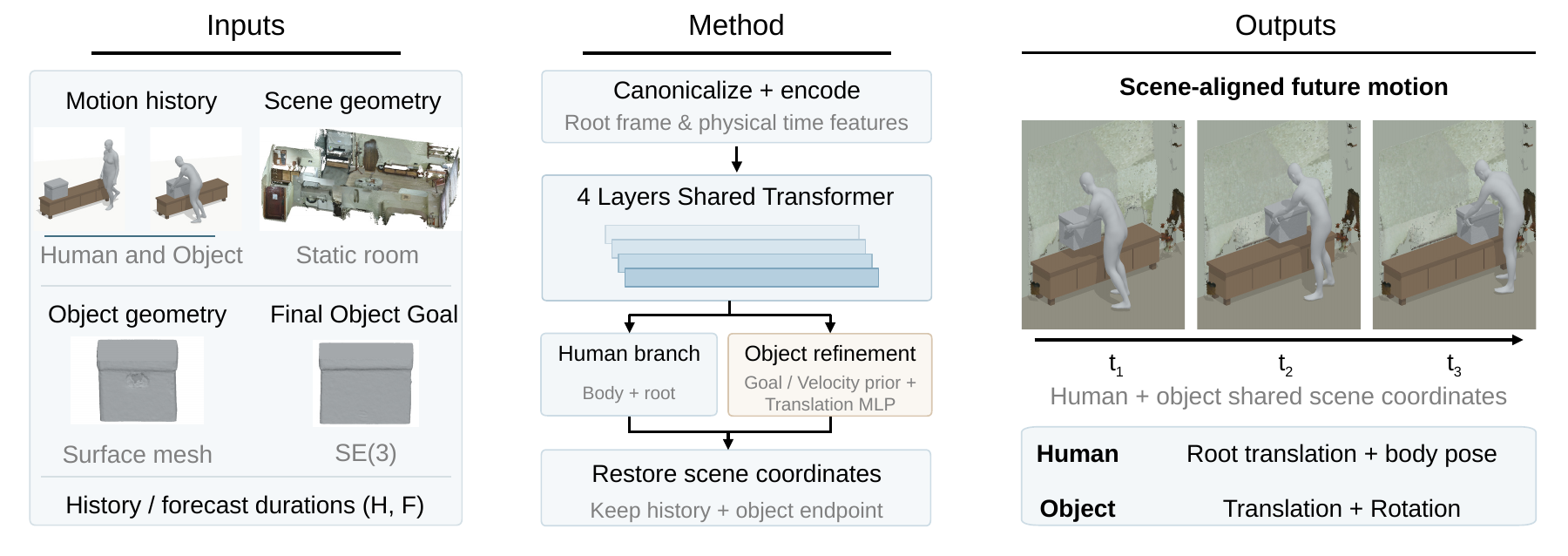}
\caption{\METHOD{} predicts joint motion from history, scene/object geometry,
the object goal, and query times. An endpoint-preserving head refines object
translation before both trajectories are restored to the metric scene frame.}
\label{fig:benchmark_pipeline}
\vspace{-8pt}
\end{figure}

\begingroup
\setlength{\parskip}{3pt}
\section{\BENCHMARK}
\label{sec:benchmark}

Existing HOI prediction and synthesis protocols provide different amounts
and types of future information, making their results difficult to compare
under a common task~\citep{xu2023interdiff,li2023omomo}.
To provide a consistent setting, we introduce \textbf{\BENCHMARK{}}
on \DATASET{}. Given observed human-object motion, scene and object
geometry, and a target object pose, the benchmark tests whether a model
can predict their coordinated motion toward that target across varying
observation and forecast durations. 
We evaluate both world-space
trajectory accuracy and human-object coordination, and propose
\textbf{\METHOD{}} as a reference method.

\subsection{Benchmark definition and protocol}
\label{sec:benchmark_task}
We formalize each interaction as a sequence of human and object states.
At time $t$, the state is
$
X_t=(q_t,\Theta_t,p_t,R_t),
$
where $q_t$ is the human-root translation, $\Theta_t$ contains the scene-frame root orientation and 21 SMPL-X body rotations, and $(p_t,R_t)$ is the object translation and rotation.

Let $t=0$ denote the last observed time. For a history duration $H$ and forecast duration $F$, let $\mathcal Q_H\subset[-H,0]$ and $\mathcal Q_F\subset(0,F]$ denote the sampled history and future timestamps. The task is
\begin{equation}
\widehat X_{\mathcal Q_F}=f_\theta\!\left(
X_{\mathcal Q_H},\mathcal S,\mathcal O,g,
\mathcal Q_H,\mathcal Q_F\right),
\qquad
g=(p_F,R_F),
\label{eq:benchmark_task}
\end{equation}
where $\mathcal S$ and $\mathcal O$ denote the static scene and object geometry, and $g$ is the terminal object pose obtained from the reconstructed trajectory. At inference time, $g$ is provided as a condition. The model predicts the complete future human motion and the intermediate object trajectory, with the object pose at $t=F$ constrained to match $g$.

\vspace{-1ex}
\paragraph{Data and partition.}
We extract 135-frame source windows from quality-screened dynamic-object clips with paired human and object trajectories and known object meshes. 
Each window spans approximately 4.5 s at 30 Hz. 
This yields 26,404 training, 5,527 validation, and 6,948 testing windows. 
We partition the data by recording so that overlapping windows from the same recording never cross splits, and compute all normalization statistics using training data only. Appendix~\ref{app:benchmark} details the window-extraction and trajectory-validity requirements.

\vspace{-1ex}
\paragraph{Flexible-time protocol.}
We train a single model across variable history and forecast durations. 
For each source window, we sample the last-observation boundary $t=0$ together with continuous durations $(H,F)$ satisfying $H+F\leq4$ s; 
the remaining portion of the 4.5-second window allows the boundary to shift across training examples. 
Motion states are sampled at the requested timestamps using linear interpolation for translations and spherical linear interpolation for rotations. 
At evaluation, the same trained model is applied without retraining to 11 fixed duration pairs at 16 Hz. 
Training excludes small neighborhoods around $(H,F)=(0.85,1.6)$ s and $(1.5,0.8)$ s, so these pairs test generalization to held-out duration combinations. 
Checkpoint selection uses six validation pairs outside the excluded neighborhoods. Appendix~\ref{app:benchmark} specifies the query counts, excluded ranges, and validation grid; Section~\ref{sec:benchmark_results} reports the evaluation protocol and results.

\subsection{\METHOD: shared prediction with object refinement}
\label{sec:benchmark_model}

We further propose \METHOD{} as reference model for \BENCHMARK{}. It couples human and object predictions through
shared temporal features, then refines object translation while preserving
the supplied terminal pose (Fig.~\ref{fig:benchmark_pipeline}).

To represent motion and geometry consistently, we express the observed
states, target pose, and scene geometry in a local frame centered at the
last observed human root and aligned with its yaw. This removes arbitrary
global translation and heading while preserving their relative geometry.
Separate point encoders aggregate 256 samples from the object mesh in its
local frame and the 512 scene points nearest the last observed root.

A four-layer, width-192 Transformer processes observed-state and
future-query tokens with continuous-time and history/future/goal role encodings.
It also receives the last observed state, terminal object pose, history
and forecast durations, object geometry, and local scene geometry.
Its shared temporal features predict human and object residuals relative
to simple reference motions: the human state is held at its last observed
value, while the object pose is interpolated toward the terminal goal.
We represent body rotations in continuous 6D form~\citep{zhou2019rotation6d}.

We train the shared predictor for 120 epochs with weighted state-reconstruction
and physical-time velocity losses. To supervise human-object coordination, we also predict
the distances from both wrists to the object surface, computing wrist
positions through SMPL-X forward kinematics with a neutral body shape.
We then freeze the shared predictor and train only the object-translation
refinement head for up to 40 additional epochs, preserving the predicted
human motion.
Architecture and loss details are provided in
Appendix~\ref{app:benchmark_implementation}.

\label{sec:benchmark:end}
\par
\endgroup
\allowbreak
\begin{table}[!t]
\centering
\setlength{\abovecaptionskip}{0pt}
\caption{Accuracy on the 30-minute MoCap subset. OSX~\citep{lin2023osx} and FoundationPose~\citep{wen2024foundationpose}
are the initial human and object estimates. Each ``w/o'' row removes a
component from the full pipeline: $E_D$ is the depth-based human surface
loss, and $E_{\Delta^2 x_t}$ denotes all human temporal terms in
Eq.~\ref{eq:human_fitting}; object refinement is defined in
Eq.~\ref{eq:object_refinement}, and tracking safeguards are described under
Object motion in Sec.~\ref{sec:dataset_reconstruction}. All configurations
use the same calibration and valid samples for each applicable metric.
Dashes denote inapplicable or unreported metrics.}
\label{tab:mocap_accuracy}
\begingroup
\fontsize{8}{9.5}\selectfont
\setlength{\tabcolsep}{0pt}
\renewcommand{\arraystretch}{1.00}
\setlength{\aboverulesep}{1pt}
\setlength{\belowrulesep}{1pt}
\begin{tabular}{@{}
>{\raggedright\arraybackslash}m{0.27\linewidth}
>{\centering\arraybackslash}m{0.14\linewidth}
>{\centering\arraybackslash}m{0.16\linewidth}
>{\centering\arraybackslash}m{0.15\linewidth}
>{\centering\arraybackslash}m{0.115\linewidth}
>{\centering\arraybackslash}m{0.165\linewidth}
@{}}
\toprule
\textbf{Pipeline configuration}
& \mbox{\textbf{G-MPJPE}/mm$\downarrow$}
& \mbox{\textbf{RA-MPJPE}/mm$\downarrow$}
& \mbox{\textbf{Obj. trans.}/mm$\downarrow$}
& \mbox{\textbf{Obj. rot.}/$^\circ\downarrow$}
& \mbox{\textbf{HOI-relative}/mm$\downarrow$} \\
\midrule
OSX
& 195.35 & 168.84 & \textendash & \textendash & \textendash \\
FoundationPose
& \textendash & \textendash & 43.0 & 11.6 & \textendash \\
\midrule
w/o $E_D$
& 73.6 & 50.0 & 28.7 & 7.5 & 55.2 \\
w/o $E_{\Delta^2 x_t}$
& 79.3 & 55.4 & 28.7 & 7.5 & 61.5 \\
w/o object refinement
& 68.3 & 47.6 & 32.2 & 8.9 & 54.8 \\
w/o tracking safeguards
& 68.3 & 47.6 & 48.7 & 13.2 & 69.6 \\
\midrule
\rowcolor{d2hoirow}
\textbf{Full pipeline}
& \textbf{68.3} & \textbf{47.6} & \textbf{28.7}
& \textbf{7.5} & \textbf{51.6} \\
\bottomrule
\end{tabular}
\endgroup
\vspace{-12pt}
\end{table}

\begingroup
\setlength{\parskip}{2pt}
\vspace{-6pt}
\section{Experiments}
\label{sec:experiments}
\vspace{-3pt}

We evaluate three questions: 
(1) how accurately the pipeline reconstructs scene-aligned human-object motion and which components contribute to its performance; 
(2) whether a single model can jointly predict human and object motion across different observation and forecast durations; 
and (3) whether \DATASET{} provides useful supervision beyond \BENCHMARK{} through cross-dataset transfer and aligned language and scene context.

\vspace{-3pt}
\subsection{Reconstruction Quality and Component Ablations}
\label{sec:pipeline_experiments}
\label{sec:mocap_validation}
\vspace{-2pt}

We assess 3D accuracy on approximately 30 minutes of synchronized
RGB-D and motion-capture (MoCap) interactions from two participants.
Fixed space-time calibration aligns matched joints and object-mesh
origins. Global mean per-joint position error (G-MPJPE) uses scene
coordinates without per-frame alignment, whereas root-aligned MPJPE
(RA-MPJPE) removes only root translation. We also report object
translation and rotation errors and object-local wrist error
(HOI-relative). All configurations use the same valid samples for
each applicable metric (Appendix~\ref{app:metrics}).

OSX~\citep{lin2023osx} and FoundationPose~\citep{wen2024foundationpose}
provide the initial human and object estimates in our pipeline
(Sec.~\ref{sec:dataset}). These estimates have substantial 3D errors
(Tab.~\ref{tab:mocap_accuracy}): OSX yields 195.35 mm G-MPJPE and
168.84 mm RA-MPJPE, while FoundationPose yields 43.0 mm object
translation error and $11.6^\circ$ rotation error. The full pipeline
reduces the corresponding errors to 68.3 mm, 47.6 mm, 28.7 mm, and
$7.5^\circ$.

\begingroup
\setlength{\intextsep}{0pt}
\setlength{\abovecaptionskip}{0.5pt}
\begin{wrapfigure}{r}{0.50\textwidth}
\centering
\includegraphics[width=\linewidth,trim=14bp 4bp 0bp 0bp,clip]
{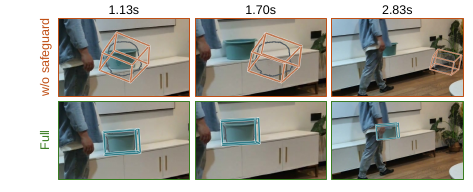}
\caption{Tracking over 1--3 s without (top) and with (bottom) safeguards.}
\label{fig:object_tracking_ablation}
\end{wrapfigure}
Human pose fitting contributes substantially to this improvement.
Removing the temporal terms $E_{\Delta^2 x_t}$ increases G-MPJPE and
RA-MPJPE by 11.0 and 7.8 mm, respectively; removing the depth-based
surface loss $E_D$ increases them by 5.3 and 2.4 mm. The two ablations
also increase HOI-relative error from 51.6 mm to 61.5 and 55.2 mm,
respectively, indicating that human fitting improves both joint
accuracy and human-object alignment.
Object trajectory refinement reduces translation and rotation errors
from 32.2 mm and $8.9^\circ$ to 28.7 mm and $7.5^\circ$. Object tracking
safeguards have a larger effect: they reduce these errors from 48.7 mm
and $13.2^\circ$ to 28.7 mm and $7.5^\circ$, and reduce HOI-relative
error from 69.6 to 51.6 mm. Fig.~\ref{fig:object_tracking_ablation}
illustrates the resulting improvement in tracking continuity and mask
alignment.
\par
\ifnum\value{WF@wrappedlines}>1
\vspace{\dimexpr\value{WF@wrappedlines}\baselineskip-\baselineskip\relax}
\fi
\WFclear
\endgroup

Since different components in our data pipeline
affect different aspects of tracking quality, we provide more extensive
ablations of these components and their specific effects in
Appendix~\ref{app:pipeline_ablations}.

\label{sec:mocap_validation:end}
\par
\endgroup

\begingroup
\setlength{\intextsep}{6pt}
\begin{table}[!t]
\centering
\setlength{\abovecaptionskip}{0pt}
\caption{\BENCHMARK{} results:
6,948 testing windows, equal-weight averages over 11 pairs, and representative
short, training-excluded, and long forecasts.}
\label{tab:benchmark_main}
\begingroup
\scriptsize
\setlength{\tabcolsep}{0.8pt}
\renewcommand{\arraystretch}{1.14}
\resizebox{\linewidth}{!}{%
\begin{tabular}{@{}l*{12}{r}@{}}
\toprule
Method & \multicolumn{3}{c}{All 11 pairs} & \multicolumn{3}{c}{$H=\text{3.0}$ s, $F=\text{0.5}$ s} & \multicolumn{3}{c}{$H=\text{0.85}$ s, $F=\text{1.6}$ s} & \multicolumn{3}{c}{$H=\text{0.25}$ s, $F=\text{3.75}$ s} \\
\cmidrule(lr){2-4}\cmidrule(lr){5-7}\cmidrule(lr){8-10}\cmidrule(l){11-13}
& Root/mm$\downarrow$ & Obj./mm$\downarrow$ &  F1(\%)$\uparrow$ & Root/mm$\downarrow$ & Obj./mm$\downarrow$ &  F1(\%)$\uparrow$ & Root/mm$\downarrow$ & Obj./mm$\downarrow$ &  F1(\%)$\uparrow$ & Root/mm$\downarrow$ & Obj./mm$\downarrow$ &  F1(\%)$\uparrow$ \\
\midrule
CHOIS & 244.7\textpm5.4 & 65.8\textpm0.04 & 68.2\textpm0.6 & 126.4\textpm1.9 & 21.8\textpm0.06 & 80.9\textpm0.9 & 248.0\textpm3.4 & 69.9\textpm0.05 & 65.2\textpm0.2 & 457.7\textpm9.6 & 143.5\textpm0.02 & 48.6\textpm0.6 \\
DecHOI & 190.8\textpm0.5 & 65.8\textpm0.02 & 72.1\textpm0.2 & 98.5\textpm0.2 & 21.8\textpm0.00 & 83.1\textpm0.5 & 188.6\textpm0.8 & 69.7\textpm0.02 & 70.0\textpm0.2 & 362.2\textpm1.3 & 143.6\textpm0.01 & 54.4\textpm0.6 \\
InterDiff & 265.8\textpm0.6 & 83.5\textpm0.33 & 66.7\textpm0.3 & 137.2\textpm4.2 & 42.8\textpm2.21 & 79.8\textpm0.6 & 263.3\textpm1.8 & 87.8\textpm0.30 & 63.8\textpm0.9 & 507.6\textpm2.4 & 158.8\textpm0.56 & 45.7\textpm0.1 \\
OMOMO & 322.3\textpm4.8 & 65.4\textpm0.00 & 60.9\textpm0.6 & 149.1\textpm1.4 & 21.4\textpm0.00 & 78.0\textpm0.9 & 334.4\textpm6.1 & 69.3\textpm0.00 & 55.6\textpm1.1 & 628.5\textpm25.1 & 143.1\textpm0.00 & 37.7\textpm0.7 \\
\midrule
\rowcolor{d2hoirow}
\textbf{\METHOD} & \textbf{187.2\textpm3.3} & \textbf{61.0\textpm0.22} & \textbf{74.9\textpm0.4} & \textbf{96.3\textpm3.0} & \textbf{19.7\textpm0.02} & \textbf{85.3\textpm0.4} & \textbf{186.9\textpm3.2} & \textbf{64.8\textpm0.22} & \textbf{73.1\textpm0.4} & \textbf{355.0\textpm4.1} & \textbf{132.9\textpm0.66} & \textbf{58.1\textpm0.7} \\
\bottomrule
\end{tabular}
}%
\endgroup
\vspace{-8pt}
\end{table}
\endgroup

\begingroup
\setlength{\parskip}{2pt}
\vspace{-3pt}
\subsection{Evaluation on \BENCHMARK{}}
\label{sec:benchmark_results}
\vspace{-2pt}

We compare \METHOD{} with four task-adapted baselines, including
CHOIS~\citep{li2024chois}, DecHOI~\citep{jung2026dechoi},
InterDiff~\citep{xu2023interdiff}, and OMOMO~\citep{li2023omomo}, on \BENCHMARK{}. 
All methods share the same partition and state/geometry/goal/duration interface and are trained for 120 epochs. \METHOD{} adds up to 40 refinement epochs
(details in Appendix~\ref{app:benchmark_implementation}).

\vspace{-1ex}
\paragraph{Evaluation metrics.}
\label{sec:benchmark_evaluation}
We compare methods across history and forecast duration pairs using three
metrics. Human-root and object translation errors measure trajectory
accuracy against ground truth. Proximity F1 measures whether the predicted wrists are near the object at the correct interactoin frames:
for each wrist, we classify its distance to the object surface as near
when it is below 20 cm, then compare these states with the ground truth. More details can be found in Appendix~\ref{app:benchmark_body_metrics}.

\vspace{-1ex}
\paragraph{Benchmark comparison and duration generalization.} 
Tab.~\ref{tab:benchmark_main} shows that \METHOD{} leads on all three
metrics in the aggregate and in every duration setting shown. Averaged
equally over the 11 pairs, it reduces root error by 1.9\% relative to
DecHOI (187.2 vs.\ 190.8 mm) and object error by 6.7\% relative to the
best baseline (61.0 vs.\ 65.4 mm), while improving proximity F1 by
2.8 percentage points (74.9\% vs.\ 72.1\%). The advantage persists for
the long forecast ($F=3.75$ s), where \METHOD{} achieves 355.0 mm root
error, 132.9 mm object error, and 58.1\% proximity F1. These results indicate that \METHOD{} improves both trajectory accuracy
and human-object coordination across the evaluated forecast horizons.

\label{sec:benchmark_results:end}

\begingroup
\setlength{\tabcolsep}{2pt}
\begin{table}[!t]
\centering
\setlength{\abovecaptionskip}{0pt}
\caption{Transfer to BEHAVE under matched 6k- and 12k-step training
budgets. For each budget, the control trains entirely on BEHAVE, while
the transfer variant trains for the first 3k steps on \DATASET{} and
the remaining steps on BEHAVE. All models are evaluated on BEHAVE.
InterDiff reports eight DDIM sampling seeds and HOIMotion three runs;
values are means $\pm$ sample SD. Bold denotes the better mean within
each comparison.}
\label{tab:external_transfer}
\begingroup
\small
\setlength{\tabcolsep}{3pt}
\renewcommand{\arraystretch}{1.0}
\setlength{\aboverulesep}{1pt}
\setlength{\belowrulesep}{1pt}
\begin{tabular*}{\linewidth}{@{\extracolsep{\fill}}llr>{\columncolor{d2hoirow}}rr>{\columncolor{d2hoirow}[\tabcolsep][0pt]}r@{}}
\toprule
& & \multicolumn{2}{c}{6k total steps}
  & \multicolumn{2}{c}{12k total steps} \\
\cmidrule(lr){3-4}\cmidrule(l){5-6}
Method & Metric (mm$\downarrow$)
& BEHAVE only & +\DATASET{} pretrain
& BEHAVE only & +\DATASET{} pretrain \\
\midrule
InterDiff & Object ADE
& 157.39\textpm4.76 & \textbf{141.32\textpm4.25}
& 155.24\textpm4.32 & \textbf{139.36\textpm4.68} \\
HOIMotion & Body MPJPE
& 67.93\textpm0.34 & \textbf{56.59\textpm0.47}
& 67.73\textpm0.18 & \textbf{55.43\textpm0.49} \\
\bottomrule
\end{tabular*}
\endgroup
\vspace{-8pt}
\end{table}
\endgroup

\vspace{-3pt}
\subsection{Cross-dataset transfer}
\label{sec:external_validation}
\vspace{-2pt}

We test whether \DATASET{} benefits existing HOI methods on
BEHAVE~\citep{bhatnagar2022behave}. For InterDiff~\citep{xu2023interdiff}
and HOIMotion~\citep{hu2024hoimotion}, we compare training entirely on
BEHAVE with training for the first 3k steps on \DATASET{} and the
remaining steps on BEHAVE, holding the total budget at either 6k or
12k steps. All models are evaluated on BEHAVE.
As shown in Tab.~\ref{tab:external_transfer}, pretraining reduces
InterDiff object ADE from 157.39 to 141.32 mm at 6k steps (10.21\%)
and from 155.24 to 139.36 mm at 12k steps (10.23\%). It also reduces
HOIMotion body MPJPE from 67.93 to 56.59 mm (16.69\%) and from
67.73 to 55.43 mm (18.16\%), respectively. 
These gains across both methods and training budgets show that
\DATASET{} provides useful pretraining data for HOI modeling beyond
\BENCHMARK{}.

\label{sec:external_validation:end}
\par
\endgroup

\begingroup
\setlength{\parskip}{4pt}
\vspace{-5pt}
\section{Conclusion}
\label{sec:conclusion}
\vspace{-3pt}

We present \DATASET{}, a scene-grounded HOI dataset with 86 hours of
recordings, 18,429 annotated clips, and 16 indoor rooms. Captured with a
single RGB-D camera, it aligns full-body and rigid-object trajectories
with scene geometry and segment-level language. \BENCHMARK{} evaluates
goal-conditioned joint prediction across observation and forecast durations,
with \METHOD{} as a reference model. MoCap validation supports reconstruction
quality, while benchmark comparisons and external-model evaluations demonstrate
its value for human-object motion modeling.

\vspace{-1ex}
\paragraph{Limitations.}
The data covers indoor interactions with rigid objects, while the benchmark
provides the terminal object pose and limits $H+F$ to 4 s for manageable training. Future work could
extend coverage to articulated and deformable objects and longer interactions,
infer goals from observations, and evaluate contact, support, and collisions
more explicitly.
\label{sec:conclusion:end}
\par
\endgroup

\allowbreak

\subsection*{AI Use Statement}

In this work, we used generative AI tools for grammar checking, LaTeX formatting, and format checks. AI-assisted compiled manuscript was inspected for formatting and layout issues.
We take responsibility for the final content of this work, including text, claims, and artifacts produced with the aid of generative AI.\allowbreak
\subsection*{Ethics Statement}

\DATASET{} was collected with the consent of all participants for recording,
public data release, and the publication of identifiable images in the paper
and dataset. Third-party assets remain subject to their
respective licenses. The resource supports research in human-object
interaction, scene-aware motion modeling, and multimodal learning. As RGB-D
observations and reconstructed indoor scenes can retain identifying
information, downstream users should respect the agreed scope of consent
and avoid re-identification or non-consensual surveillance. Responsible
reuse should also consider the coverage of the collected participants and
environments when deploying models beyond the dataset.
\allowbreak
\subsection*{Reproducibility Statement}

Sections~\ref{sec:dataset}--\ref{sec:experiments} describe dataset construction,
\BENCHMARK, and the experimental protocols.
Appendices~\ref{app:dataset} and~\ref{app:pipeline} document data formats,
coordinate conventions, reconstruction objectives, and implementation settings.
Appendix~\ref{app:validation} specifies ablation cohorts, metrics, and aggregation.
Appendix~\ref{app:benchmark} provides window construction, recording-level
partitions, temporal queries, model architectures, training settings, random
seeds, and baseline adaptations. External-model protocols are detailed in
Appendix~\ref{app:external_validation}; synchronized MoCap calibration and
accuracy metrics are described in Section~\ref{sec:mocap_validation}.
An anonymized code repository is available at
\url{https://anonymous.4open.science/r/iclr2027-anonymous-code-1DFB}.
The full dataset and code will be publicly released upon acceptance.
\allowbreak

\bibliography{references}

@inproceedings{tevet2023mdm,
  author = {Tevet, Guy and Raab, Sigal and Gordon, Brian and Shafir, Yoni and Cohen-Or, Daniel and Bermano, Amit Haim},
  title = {Human Motion Diffusion Model},
  booktitle = {International Conference on Learning Representations},
  year = {2023},
  url = {https://openreview.net/forum?id=SJ1kSyO2jwu},
  sourceurl = {https://guytevet.github.io/mdm-page/}
}

@inproceedings{karunratanakul2023gmd,
  author = {Karunratanakul, Korrawe and Preechakul, Konpat and Suwajanakorn, Supasorn and Tang, Siyu},
  title = {Guided Motion Diffusion for Controllable Human Motion Synthesis},
  booktitle = {Proceedings of the IEEE/CVF International Conference on Computer Vision},
  pages = {2151--2162},
  year = {2023},
  sourceurl = {https://openaccess.thecvf.com/content/ICCV2023/html/Karunratanakul_Guided_Motion_Diffusion_for_Controllable_Human_Motion_Synthesis_ICCV_2023_paper.html}
}

@inproceedings{he2025omnih2o,
  author = {He, Tairan and Luo, Zhengyi and He, Xialin and Xiao, Wenli and Zhang, Chong and Zhang, Weinan and Kitani, Kris M. and Liu, Changliu and Shi, Guanya},
  title = {{OmniH2O}: Universal and Dexterous Human-to-Humanoid Whole-Body Teleoperation and Learning},
  booktitle = {Proceedings of The 8th Conference on Robot Learning},
  series = {Proceedings of Machine Learning Research},
  volume = {270},
  pages = {1516--1540},
  year = {2025},
  url = {https://proceedings.mlr.press/v270/he25b.html}
}

@inproceedings{xu2025intermimic,
  author = {Xu, Sirui and Ling, Hung Yu and Wang, Yu-Xiong and Gui, Liang-Yan},
  title = {{InterMimic}: Towards Universal Whole-Body Control for Physics-Based Human-Object Interactions},
  booktitle = {Proceedings of the IEEE/CVF Conference on Computer Vision and Pattern Recognition},
  pages = {12266--12277},
  year = {2025},
  sourceurl = {https://openaccess.thecvf.com/content/CVPR2025/html/Xu_InterMimic_Towards_Universal_Whole-Body_Control_for_Physics-Based_Human-Object_Interactions_CVPR_2025_paper.html}
}

@inproceedings{zhang2023neuraldome,
  author = {Zhang, Juze and Luo, Haimin and Yang, Hongdi and Xu, Xinru and Wu, Qianyang and Shi, Ye and Yu, Jingyi and Xu, Lan and Wang, Jingya},
  title = {{NeuralDome}: A Neural Modeling Pipeline on Multi-View Human-Object Interactions},
  booktitle = {Proceedings of the IEEE/CVF Conference on Computer Vision and Pattern Recognition},
  pages = {8834--8845},
  year = {2023},
  sourceurl = {https://openaccess.thecvf.com/content/CVPR2023/html/Zhang_NeuralDome_A_Neural_Modeling_Pipeline_on_Multi-View_Human-Object_Interactions_CVPR_2023_paper.html}
}

@inproceedings{hassan2019prox,
  author = {Hassan, Mohamed and Choutas, Vasileios and Tzionas, Dimitrios and Black, Michael J.},
  title = {Resolving {3D} Human Pose Ambiguities with {3D} Scene Constraints},
  booktitle = {Proceedings of the IEEE/CVF International Conference on Computer Vision},
  pages = {2282--2292},
  year = {2019},
  sourceurl = {https://openaccess.thecvf.com/content_ICCV_2019/html/Hassan_Resolving_3D_Human_Pose_Ambiguities_With_3D_Scene_Constraints_ICCV_2019_paper.html}
}

@inproceedings{huang2022rich,
  author = {Huang, Chun-Hao P. and Yi, Hongwei and H{\"o}schle, Markus and Safroshkin, Matvey and Alexiadis, Tsvetelina and Polikovsky, Senya and Scharstein, Daniel and Black, Michael J.},
  title = {Capturing and Inferring Dense Full-Body Human-Scene Contact},
  booktitle = {Proceedings of the IEEE/CVF Conference on Computer Vision and Pattern Recognition},
  year = {2022},
  sourceurl = {https://openaccess.thecvf.com/content/CVPR2022/papers/Huang_Capturing_and_Inferring_Dense_Full-Body_Human-Scene_Contact_CVPR_2022_paper.pdf}
}

@inproceedings{liu2022hoi4d,
  author = {Liu, Yunze and Liu, Yun and Jiang, Che and Lyu, Kangbo and Wan, Weikang and Shen, Hao and Liang, Boqiang and Fu, Zhoujie and Wang, He and Yi, Li},
  title = {{HOI4D}: A {4D} Egocentric Dataset for Category-Level Human-Object Interaction},
  booktitle = {Proceedings of the IEEE/CVF Conference on Computer Vision and Pattern Recognition},
  pages = {21013--21022},
  year = {2022},
  sourceurl = {https://openaccess.thecvf.com/content/CVPR2022/html/Liu_HOI4D_A_4D_Egocentric_Dataset_for_Category-Level_Human-Object_Interaction_CVPR_2022_paper.html}
}

@inproceedings{fan2023arctic,
  author = {Fan, Zicong and Taheri, Omid and Tzionas, Dimitrios and Kocabas, Muhammed and Kaufmann, Manuel and Black, Michael J. and Hilliges, Otmar},
  title = {{ARCTIC}: A Dataset for Dexterous Bimanual Hand-Object Manipulation},
  booktitle = {Proceedings of the IEEE/CVF Conference on Computer Vision and Pattern Recognition},
  year = {2023},
  url = {https://arctic.is.tue.mpg.de/}
}

@inproceedings{taheri2020grab,
  author = {Taheri, Omid and Ghorbani, Nima and Black, Michael J. and Tzionas, Dimitrios},
  title = {{GRAB}: A Dataset of Whole-Body Human Grasping of Objects},
  booktitle = {European Conference on Computer Vision},
  year = {2020},
  url = {https://grab.is.tue.mpg.de/}
}

@inproceedings{bhatnagar2022behave,
  author = {Bhatnagar, Bharat Lal and Xie, Xianghui and Petrov, Ilya A. and Sminchisescu, Cristian and Theobalt, Christian and Pons-Moll, Gerard},
  title = {{BEHAVE}: Dataset and Method for Tracking Human Object Interactions},
  booktitle = {Proceedings of the IEEE/CVF Conference on Computer Vision and Pattern Recognition},
  pages = {15935--15946},
  year = {2022},
  sourceurl = {https://openaccess.thecvf.com/content/CVPR2022/html/Bhatnagar_BEHAVE_Dataset_and_Method_for_Tracking_Human_Object_Interactions_CVPR_2022_paper.html}
}

@inproceedings{huang2022intercap,
  author = {Huang, Yinghao and Taheri, Omid and Black, Michael J. and Tzionas, Dimitrios},
  title = {{InterCap}: Joint Markerless {3D} Tracking of Humans and Objects in Interaction},
  booktitle = {Pattern Recognition: 44th DAGM German Conference},
  series = {Lecture Notes in Computer Science},
  volume = {13485},
  pages = {281--299},
  year = {2022},
  url = {https://intercap.is.tue.mpg.de/}
}

@inproceedings{zhang2024hoim3,
  author = {Zhang, Juze and Zhang, Jingyan and Song, Zining and Shi, Zhanhe and Zhao, Chengfeng and Shi, Ye and Yu, Jingyi and Xu, Lan and Wang, Jingya},
  title = {{HOI-M$^3$}: Capture Multiple Humans and Objects Interaction within Contextual Environment},
  booktitle = {Proceedings of the IEEE/CVF Conference on Computer Vision and Pattern Recognition},
  year = {2024},
  url = {https://juzezhang.github.io/HOIM3_ProjectPage/}
}

@inproceedings{kim2025parahome,
  author = {Kim, Jeonghwan and Kim, Jisoo and Na, Jeonghyeon and Joo, Hanbyul},
  title = {{ParaHome}: Parameterizing Everyday Home Activities Towards {3D} Generative Modeling of Human-Object Interactions},
  booktitle = {Proceedings of the IEEE/CVF Conference on Computer Vision and Pattern Recognition},
  year = {2025},
  url = {https://jlogkim.github.io/parahome/}
}

@inproceedings{lu2025humoto,
  author = {Lu, Jiaxin and Huang, Chun-Hao Paul and Bhattacharya, Uttaran and Huang, Qixing and Zhou, Yi},
  title = {{HUMOTO}: A {4D} Dataset of Mocap Human Object Interactions},
  booktitle = {Proceedings of the IEEE/CVF International Conference on Computer Vision},
  pages = {10886--10897},
  year = {2025},
  sourceurl = {https://openaccess.thecvf.com/content/ICCV2025/html/Lu_HUMOTO_A_4D_Dataset_of_Mocap_Human_Object_Interactions_ICCV_2025_paper.html}
}

@inproceedings{schonberger2016sfm,
  author = {Sch{\"o}nberger, Johannes L. and Frahm, Jan-Michael},
  title = {Structure-from-Motion Revisited},
  booktitle = {Proceedings of the IEEE Conference on Computer Vision and Pattern Recognition},
  year = {2016},
  sourceurl = {https://openaccess.thecvf.com/content_cvpr_2016/papers/Schonberger_Structure-From-Motion_Revisited_CVPR_2016_paper.pdf}
}

@inproceedings{sarlin2019hloc,
  author = {Sarlin, Paul-Edouard and Cadena, Cesar and Siegwart, Roland and Dymczyk, Marcin},
  title = {From Coarse to Fine: Robust Hierarchical Localization at Large Scale},
  booktitle = {Proceedings of the IEEE/CVF Conference on Computer Vision and Pattern Recognition},
  year = {2019},
  sourceurl = {https://openaccess.thecvf.com/content_CVPR_2019/papers/Sarlin_From_Coarse_to_Fine_Robust_Hierarchical_Localization_at_Large_Scale_CVPR_2019_paper.pdf}
}

@inproceedings{pavlakos2019smplx,
  author = {Pavlakos, Georgios and Choutas, Vasileios and Ghorbani, Nima and Bolkart, Timo and Osman, Ahmed A. A. and Tzionas, Dimitrios and Black, Michael J.},
  title = {Expressive Body Capture: {3D} Hands, Face, and Body from a Single Image},
  booktitle = {Proceedings of the IEEE/CVF Conference on Computer Vision and Pattern Recognition},
  pages = {10975--10985},
  year = {2019},
  url = {https://smpl-x.is.tue.mpg.de/}
}

@inproceedings{lin2023osx,
  author = {Lin, Jing and Zeng, Ailing and Wang, Haoqian and Zhang, Lei and Li, Yu},
  title = {One-Stage {3D} Whole-Body Mesh Recovery with Component Aware Transformer},
  booktitle = {Proceedings of the IEEE/CVF Conference on Computer Vision and Pattern Recognition},
  pages = {21159--21168},
  year = {2023},
  sourceurl = {https://openaccess.thecvf.com/content/CVPR2023/html/Lin_One-Stage_3D_Whole-Body_Mesh_Recovery_With_Component_Aware_Transformer_CVPR_2023_paper.html}
}

@inproceedings{ravi2025sam2,
  author = {Ravi, Nikhila and Gabeur, Valentin and Hu, Yuan-Ting and Hu, Ronghang and Ryali, Chaitanya and Ma, Tengyu and Khedr, Haitham and R{\"a}dle, Roman and Rolland, Chloe and Gustafson, Laura and Mintun, Eric and Pan, Junting and Alwala, Kalyan Vasudev and Carion, Nicolas and Wu, Chao-Yuan and Girshick, Ross and Doll{\'a}r, Piotr and Feichtenhofer, Christoph},
  title = {{SAM 2}: Segment Anything in Images and Videos},
  booktitle = {International Conference on Learning Representations},
  year = {2025},
  url = {https://openreview.net/forum?id=Ha6RTeWMd0}
}

@inproceedings{wen2023bundlesdf,
  author = {Wen, Bowen and Tremblay, Jonathan and Blukis, Valts and Tyree, Stephen and M{\"u}ller, Thomas and Evans, Alex and Fox, Dieter and Kautz, Jan and Birchfield, Stan},
  title = {{BundleSDF}: Neural {6-DoF} Tracking and {3D} Reconstruction of Unknown Objects},
  booktitle = {Proceedings of the IEEE/CVF Conference on Computer Vision and Pattern Recognition},
  pages = {606--617},
  year = {2023},
  sourceurl = {https://openaccess.thecvf.com/content/CVPR2023/html/Wen_BundleSDF_Neural_6-DoF_Tracking_and_3D_Reconstruction_of_Unknown_Objects_CVPR_2023_paper.html}
}

@inproceedings{wen2024foundationpose,
  author = {Wen, Bowen and Yang, Wei and Kautz, Jan and Birchfield, Stan},
  title = {{FoundationPose}: Unified {6D} Pose Estimation and Tracking of Novel Objects},
  booktitle = {Proceedings of the IEEE/CVF Conference on Computer Vision and Pattern Recognition},
  pages = {17868--17879},
  year = {2024},
  sourceurl = {https://openaccess.thecvf.com/content/CVPR2024/html/Wen_FoundationPose_Unified_6D_Pose_Estimation_and_Tracking_of_Novel_Objects_CVPR_2024_paper.html}
}

@inproceedings{xu2023interdiff,
  author = {Xu, Sirui and Li, Zhengyuan and Wang, Yu-Xiong and Gui, Liang-Yan},
  title = {{InterDiff}: Generating {3D} Human-Object Interactions with Physics-Informed Diffusion},
  booktitle = {Proceedings of the IEEE/CVF International Conference on Computer Vision},
  pages = {14928--14940},
  year = {2023},
  sourceurl = {https://openaccess.thecvf.com/content/ICCV2023/html/Xu_InterDiff_Generating_3D_Human-Object_Interactions_with_Physics-Informed_Diffusion_ICCV_2023_paper.html}
}

@article{li2023omomo,
  author = {Li, Jiaman and Wu, Jiajun and Liu, C. Karen},
  title = {Object Motion Guided Human Motion Synthesis},
  journal = {ACM Transactions on Graphics},
  volume = {42},
  number = {6},
  year = {2023},
  url = {https://lijiaman.github.io/projects/omomo/}
}

@inproceedings{diller2024cghoi,
  author = {Diller, Christian and Dai, Angela},
  title = {{CG-HOI}: Contact-Guided {3D} Human-Object Interaction Generation},
  booktitle = {Proceedings of the IEEE/CVF Conference on Computer Vision and Pattern Recognition},
  pages = {19888--19901},
  year = {2024},
  sourceurl = {https://openaccess.thecvf.com/content/CVPR2024/html/Diller_CG-HOI_Contact-Guided_3D_Human-Object_Interaction_Generation_CVPR_2024_paper.html}
}

@inproceedings{li2024chois,
  author = {Li, Jiaman and Clegg, Alexander and Mottaghi, Roozbeh and Wu, Jiajun and Puig, Xavier and Liu, C. Karen},
  title = {Controllable Human-Object Interaction Synthesis},
  booktitle = {European Conference on Computer Vision},
  year = {2024},
  url = {https://lijiaman.github.io/projects/chois/}
}

@inproceedings{liu2025core4d,
  author = {Liu, Yun and Zhang, Chengwen and Xing, Ruofan and Tang, Bingda and Yang, Bowen and Yi, Li},
  title = {{CORE4D}: A {4D} Human-Object-Human Interaction Dataset for Collaborative Object {REarrangement}},
  booktitle = {Proceedings of the IEEE/CVF Conference on Computer Vision and Pattern Recognition},
  pages = {1769--1782},
  year = {2025},
  url = {https://core4d.github.io/}
}

@inproceedings{kogashi2026mmhoi,
  author = {Kogashi, Kaen and Cherian, Anoop and Kuo, Meng-Yu Jennifer},
  title = {{MMHOI}: Modeling Complex {3D} Multi-Human Multi-Object Interactions},
  booktitle = {Proceedings of the IEEE/CVF Winter Conference on Applications of Computer Vision},
  pages = {1512--1521},
  year = {2026},
  url = {https://jk-vision.github.io/mmhoi/},
  sourceurl = {https://arxiv.org/html/2510.07828v1}
}

@article{huang2024intercap,
  author={Huang, Yinghao and Taheri, Omid and Black, Michael J. and Tzionas, Dimitrios},
  title={{InterCap}: Joint Markerless {3D} Tracking of Humans and Objects in Interaction from Multi-view {RGB-D} Images},
  journal={International Journal of Computer Vision}, year={2024}, doi={10.1007/s11263-024-01984-1},
  url={https://intercap.is.tue.mpg.de/}
}

@inproceedings{jiang2023chairs,
  author={Jiang, Nan and Liu, Tengyu and Cao, Zhexuan and Cui, Jieming and Zhang, Zhiyuan and Chen, Yixin and Wang, He and Zhu, Yixin and Huang, Siyuan},
  title={Full-Body Articulated Human-Object Interaction},
  booktitle={IEEE/CVF International Conference on Computer Vision (ICCV)}, year={2023},
  sourceurl={https://openaccess.thecvf.com/content/ICCV2023/papers/Jiang_Full-Body_Articulated_Human-Object_Interaction_ICCV_2023_paper.pdf}
}

@inproceedings{zhao2024imhoi,
  author={Zhao, Chengfeng and Zhang, Juze and Du, Jiashen and Shan, Ziwei and Wang, Junye and Yu, Jingyi and Wang, Jingya and Xu, Lan},
  title={I'M {HOI}: Inertia-aware Monocular Capture of {3D} Human-Object Interactions},
  booktitle={IEEE/CVF Conference on Computer Vision and Pattern Recognition (CVPR)}, year={2024},
  url={https://afterjourney00.github.io/IM-HOI.github.io/}
}

@inproceedings{lv2024himo,
  author={Lv, Xintao and Xu, Liang and Yan, Yichao and Jin, Xin and Xu, Congsheng and Wu, Shuwen and Liu, Yifan and Li, Lincheng and Bi, Mengxiao and Zeng, Wenjun and Yang, Xiaokang},
  title={{HIMO}: A New Benchmark for Full-Body Human Interacting with Multiple Objects},
  booktitle={European Conference on Computer Vision (ECCV)}, year={2024},
  url={https://lvxintao.github.io/himo/}
}

@article{cao2026acedata,
  author={Cao, Yukang and Xie, Haozhe and Wen, Beichen and Yao, Runmao and Liu, Yinghao and Huang, Yue and Liao, Zhichao and Wang, Yunxiang and Liu, Haiheng and Tian, Xingshun and Su, Dawei and Zhuo, Long and Tao, Dacheng and Wang, Xiaogang and Pan, Liang and Liu, Ziwei},
  title={{ACE-Data-0}: Human-Centric Ambient Capture as Embodied Data Engine},
  journal={arXiv preprint arXiv:2607.28625}, year={2026},
  url={https://arxiv.org/abs/2607.28625}
}

@article{ji2026hiphi,
  author={Ji, Jiahao and Ma, Ji and Zhang, Runhan and Yu, Runyi and Wang, Wenjia and Chi, Weiheng and Peng, Qianqian and Yan, Weichao and Gu, Yongfei and Tian, Ye and Wu, Ting and Li, Longwei and Yuan, Chun and Dai, Ruoli and Han, Lei},
  title={{HiPHI}: A Large-Scale Benchmark for High-Precision Human Motion and Object-Interaction},
  journal={arXiv preprint arXiv:2608.16222}, year={2026},
  url={https://arxiv.org/abs/2608.16222}
}

@inproceedings{xu2025interact,
  author={Xu, Sirui and Li, Dongting and Zhang, Yucheng and Xu, Xiyan and Long, Qi and Wang, Ziyin and Lu, Yunzhi and Dong, Shuchang and Jiang, Hezi and Gupta, Akshat and Wang, Yu-Xiong and Gui, Liang-Yan},
  title={{InterAct}: Advancing Large-Scale Versatile {3D} Human-Object Interaction Generation},
  booktitle={IEEE/CVF Conference on Computer Vision and Pattern Recognition (CVPR)}, year={2025},
  url={https://sirui-xu.github.io/InterAct/}
}

@inproceedings{detone2018superpoint,
  author = {DeTone, Daniel and Malisiewicz, Tomasz and Rabinovich, Andrew},
  title = {{SuperPoint}: Self-Supervised Interest Point Detection and Description},
  booktitle = {Proceedings of the IEEE Conference on Computer Vision and Pattern Recognition Workshops},
  pages = {224--236},
  year = {2018},
  sourceurl = {https://openaccess.thecvf.com/content_cvpr_2018_workshops/papers/w9/DeTone_SuperPoint_Self-Supervised_Interest_CVPR_2018_paper.pdf}
}

@inproceedings{sarlin2020superglue,
  author = {Sarlin, Paul-Edouard and DeTone, Daniel and Malisiewicz, Tomasz and Rabinovich, Andrew},
  title = {{SuperGlue}: Learning Feature Matching with Graph Neural Networks},
  booktitle = {Proceedings of the IEEE/CVF Conference on Computer Vision and Pattern Recognition},
  pages = {4938--4947},
  year = {2020},
  sourceurl = {https://openaccess.thecvf.com/content_CVPR_2020/html/Sarlin_SuperGlue_Learning_Feature_Matching_With_Graph_Neural_Networks_CVPR_2020_paper.html}
}

@inproceedings{arandjelovic2016netvlad,
  author = {Arandjelovi{\'c}, Relja and Gronat, Petr and Torii, Akihiko and Pajdla, Tomas and Sivic, Josef},
  title = {{NetVLAD}: {CNN} Architecture for Weakly Supervised Place Recognition},
  booktitle = {Proceedings of the IEEE Conference on Computer Vision and Pattern Recognition},
  pages = {5297--5307},
  year = {2016},
  sourceurl = {https://openaccess.thecvf.com/content_cvpr_2016/papers/Arandjelovic_NetVLAD_CNN_Architecture_CVPR_2016_paper.pdf}
}

@inproceedings{zhang2022egobody,
  author = {Zhang, Siwei and Ma, Qianli and Zhang, Yan and Qian, Zhiyin and Kwon, Taein and Pollefeys, Marc and Bogo, Federica and Tang, Siyu},
  title = {{EgoBody}: Human Body Shape and Motion of Interacting People from Head-Mounted Devices},
  booktitle = {European Conference on Computer Vision},
  pages = {180--200},
  year = {2022},
  url = {https://sanweiliti.github.io/egobody/egobody.html}
}

@inproceedings{banerjee2025hot3d,
  author = {Banerjee, Prithviraj and Shkodrani, Sindi and Moulon, Pierre and Hampali, Shreyas and Han, Shangchen and Zhang, Fan and Zhang, Linguang and Fountain, Jade and Miller, Edward and Basol, Selen and Newcombe, Richard and Wang, Robert and Engel, Jakob Julian and Hodan, Tomas},
  title = {{HOT3D}: Hand and Object Tracking in {3D} from Egocentric Multi-View Videos},
  booktitle = {Proceedings of the IEEE/CVF Conference on Computer Vision and Pattern Recognition},
  pages = {7061--7071},
  year = {2025},
  url = {https://facebookresearch.github.io/hot3d/},
  sourceurl = {https://openaccess.thecvf.com/content/CVPR2025/html/Banerjee_HOT3D_Hand_and_Object_Tracking_in_3D_from_Egocentric_Multi-View_CVPR_2025_paper.html}
}

@inproceedings{hoque2026egodex,
  author = {Hoque, Ryan and Huang, Peide and Yoon, David J. and Sivapurapu, Mouli and Zhang, Jian},
  title = {{EgoDex}: Learning Dexterous Manipulation from Large-Scale Egocentric Video},
  booktitle = {International Conference on Learning Representations},
  year = {2026},
  url = {https://arxiv.org/abs/2505.11709},
  sourceurl = {https://github.com/apple/ml-egodex}
}

@inproceedings{jung2026dechoi,
  author = {Jung, Hwanhee and Lee, Seunggwan and Yoon, Jeongyoon and Kim, SeungHyeon and Nam, Giljoo and Huang, Qixing and Kim, Sangpil},
  title = {Decoupled Generative Modeling for Human-Object Interaction Synthesis},
  booktitle = {Proceedings of the IEEE/CVF Conference on Computer Vision and Pattern Recognition},
  pages = {2253--2263},
  year = {2026},
  url = {https://kuai-lab.github.io/cvpr2026dechoi/},
  sourceurl = {https://arxiv.org/abs/2512.19049}
}

@inproceedings{zhou2019rotation6d,
  author = {Zhou, Yi and Barnes, Connelly and Lu, Jingwan and Yang, Jimei and Li, Hao},
  title = {On the Continuity of Rotation Representations in Neural Networks},
  booktitle = {Proceedings of the IEEE/CVF Conference on Computer Vision and Pattern Recognition},
  pages = {5745--5753},
  year = {2019},
  url = {https://openaccess.thecvf.com/content_CVPR_2019/html/Zhou_On_the_Continuity_of_Rotation_Representations_in_Neural_Networks_CVPR_2019_paper.html}
}

@article{hu2024hoimotion,
  author = {Hu, Zhiming and Yin, Zheming and Haeufle, Daniel and Schmitt, Syn and Bulling, Andreas},
  title = {{HOIMotion}: Forecasting Human Motion During Human-Object Interactions Using Egocentric {3D} Object Bounding Boxes},
  journal = {IEEE Transactions on Visualization and Computer Graphics},
  volume = {30},
  number = {11},
  pages = {7375--7385},
  year = {2024},
  doi = {10.1109/TVCG.2024.3456161},
  url = {https://zhiminghu.net/hu24_hoimotion.html}
}

@inproceedings{huang2023scenediffuser,
  author = {Huang, Siyuan and Wang, Zan and Li, Puhao and Jia, Baoxiong and Liu, Tengyu and Zhu, Yixin and Liang, Wei and Zhu, Song-Chun},
  title = {Diffusion-Based Generation, Optimization, and Planning in {3D} Scenes},
  booktitle = {Proceedings of the IEEE/CVF Conference on Computer Vision and Pattern Recognition},
  pages = {16750--16761},
  year = {2023},
  sourceurl = {https://openaccess.thecvf.com/content/CVPR2023/html/Huang_Diffusion-Based_Generation_Optimization_and_Planning_in_3D_Scenes_CVPR_2023_paper.html}
}

@article{jiang2024rtmw,
  author = {Jiang, Tao and Xie, Xinchen and Li, Yining},
  title = {{RTMW}: Real-Time Multi-Person {2D} and {3D} Whole-body Pose Estimation},
  journal = {arXiv preprint arXiv:2407.08634},
  year = {2024},
  sourceurl = {https://arxiv.org/abs/2407.08634}
}

@inproceedings{petrovich2023tmr,
  author = {Petrovich, Mathis and Black, Michael J. and Varol, G{\"u}l},
  title = {{TMR}: Text-to-Motion Retrieval Using Contrastive {3D} Human Motion Synthesis},
  booktitle = {Proceedings of the IEEE/CVF International Conference on Computer Vision},
  pages = {9488--9497},
  year = {2023},
  sourceurl = {https://openaccess.thecvf.com/content/ICCV2023/html/Petrovich_TMR_Text-to-Motion_Retrieval_Using_Contrastive_3D_Human_Motion_Synthesis_ICCV_2023_paper.html}
}

@inproceedings{fang2022towards,
  author = {Fang, Qihang and Yin, Yingda and Fan, Qingnan and Xia, Fei and Dong, Siyan and Wang, Sheng and Wang, Jue and Guibas, Leonidas J. and Chen, Baoquan},
  title = {Towards Accurate Active Camera Localization},
  booktitle = {Proceedings of the European Conference on Computer Vision},
  pages = {122--139},
  year = {2022},
  doi = {10.1007/978-3-031-20080-9_8},
  sourceurl = {https://www.ecva.net/papers/eccv_2022/papers_ECCV/html/5136_ECCV_2022_paper.php}
}
\bibliographystyle{iclr2027_conference}

\clearpage
\appendix
\section*{Appendix}

\section{Dataset documentation}
\label{app:supplementary}
\label{app:dataset}

\subsection{Inventory and counting conventions}
\label{app:corpus_audit}

The corpus covers 16 distinct indoor rooms. A \emph{recording} is a continuous
RGB-D capture; an \emph{annotated clip} (or \emph{segment}) is an annotated
activity interval, and a processed \emph{sequence} stores its fitted trajectories.
\emph{Evaluation clips} are intervals selected for ablation; benchmark
\emph{windows} are resampled sequence excerpts. A \emph{room group}
(capture family) groups recordings by physical room.
Excluding segments with recording blur, severe occlusion, or unusable visual evidence
leaves approximately 86 hours of annotated activity. Acquisition screening
assesses recording usability; geometric checks and independent MoCap validation
assess trajectory consistency and 3D accuracy.
Table~\ref{tab:dataset_scale} separates all semantic activity from the
dynamic-object subset; the remainder includes human interactions with static
furniture, not merely idle footage. Each room has its own metric coordinate frame. Repeated scans, dates, and alternate scene names do not
increase the room count or establish participant identities. \DATASET{} was recorded with 15 distinct participants. Each participant
contributed the same total recording duration.

\begin{table}[H]
\centering
\caption{Corpus inventory. Hours count unions of annotation spans within
recordings; the dynamic-object subset is contained in the full corpus.}
\label{tab:dataset_scale}
\begin{tabular}{@{}lrr@{}}
\toprule
Quantity & Full semantic corpus & Dynamic-object subset \\
\midrule
Distinct indoor rooms & 16 & 16 \\
Semantic intervals & 18,429 & 10,335 \\
Annotation-covered time (h) & 85.98 & 48.64 \\
Dynamic object types & --- & 13 \\
Median interval span (s) & 15.80 & 16.03 \\
Interval-span interquartile range (s) & 12.07--20.43 & 12.47--20.35 \\
Median description length (words) & 18 & 19 \\
Description-length interquartile range & 14--22 & 15--23 \\
\bottomrule
\end{tabular}
\end{table}

\subsection{Data contract and coordinates}
\label{app:data_contract}

Interaction recordings use a single handheld RGB-D camera at 30 FPS, with
$1280\times960$ RGB and $640\times480$ depth. We retain sensor calibration
and available inertial measurement unit (IMU) readings with the room scans. In
Figure~\ref{fig:dataset_pipeline}, a language description covers frames
168--392 and three clicks on the box at frame 390 initialize its mask.
Clicks specify tracking targets, not 3D correspondences or object poses;
clip and source-frame IDs link both annotation types to the observations.

Measured RGB/depth and calibration accompany human-annotated intervals,
English descriptions, object identities, and keyframe clicks. Estimated
fields include masks, 63 confidence-bearing keypoints, camera-to-scene
transforms, SMPL-X motion (translation, 165 axis-angle pose entries, ten shape
coefficients), rigid object-to-scene transforms, visibility, and QC/provenance.
Lengths follow stored source frame IDs rather than a fixed temporal grid.
The 165 pose entries include root, body, jaw/eyes, and hands; local joint
rotations are not all world rotations.

For homogeneous points, $X^s=T^s_{c,t}X^c$ and $X^s=T^s_{o,t}X^o$ give
camera/object-to-scene mappings. Object projection is
$u=\pi(K(T^s_{c,t})^{-1}T^s_{o,t}X^o)$, where $K$ denotes camera intrinsics
and $\pi$ perspective division; scene-aligned human vertices need only
the inverse camera transform. Estimator outputs are restored to raw image
orientation before projection. Static-scan and dynamic-camera calibration
remain distinct, and resizing a depth display does not create new calibrated
measurements. Body-model versions and parameter conventions accompany exports.

\par
\Needspace{20\baselineskip}
\begingroup
\setlength{\intextsep}{5pt}
\begin{wrapfigure}{r}{0.52\textwidth}
\centering
\includegraphics[width=0.95\linewidth, height=0.31\textwidth]{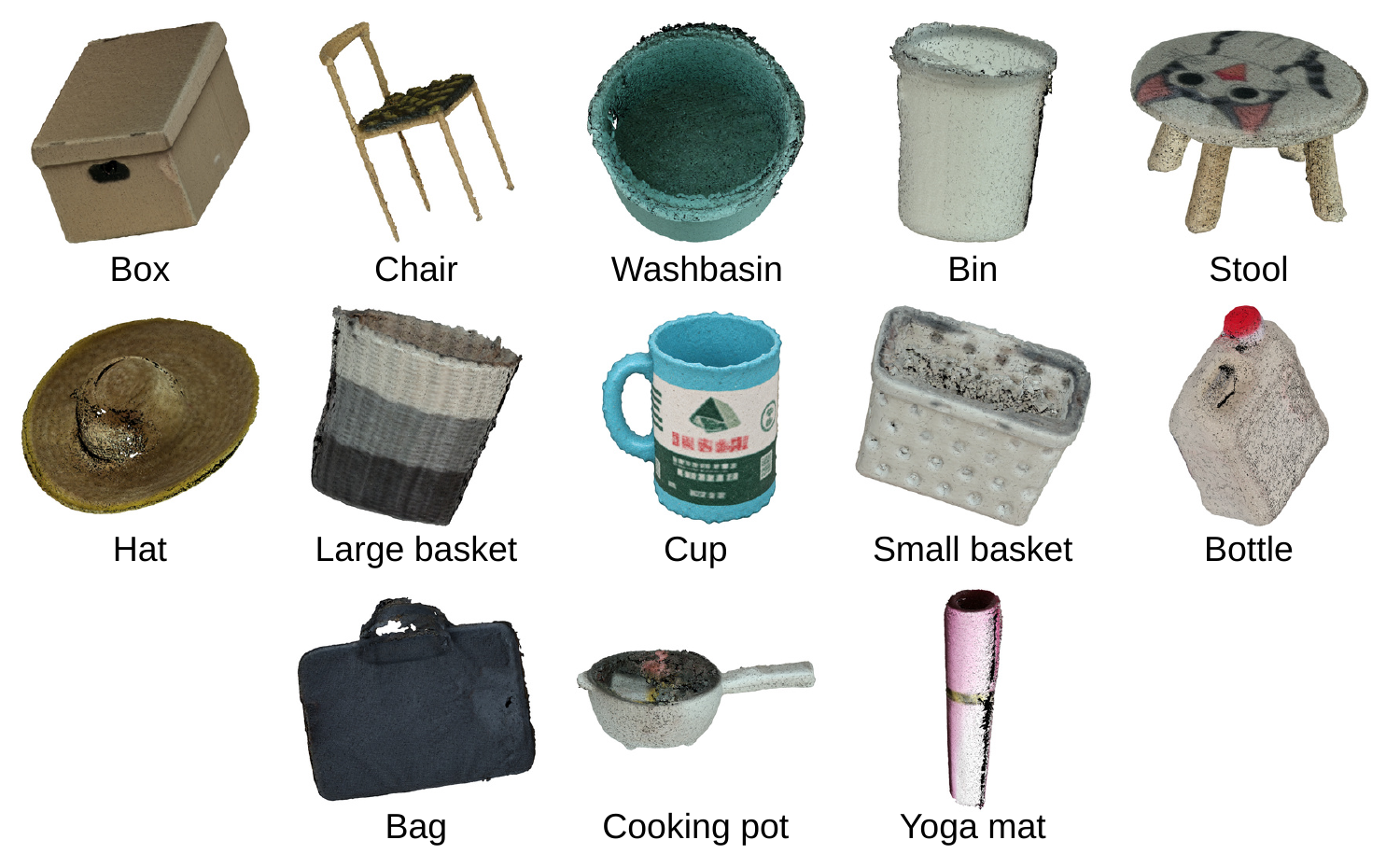}
\caption{\DATASET{}'s 13 dynamic object types, independently scaled.}
\label{fig:dataset_object_meshes}
\end{wrapfigure}
\vspace{-1ex}
\paragraph{Object mesh acquisition.}
We reconstruct the meshes for the 13 object types in
Figure~\ref{fig:dataset_object_meshes} from dedicated RGB-D scans, separate
from the interaction recordings, using BundleSDF~\citep{wen2023bundlesdf}.
Calibrated color images, depth maps, and object foreground masks support
multi-view pose tracking and optimization of a neural signed distance field
(SDF), whose zero level set defines the object surface. Global refinement
jointly optimizes this field and the selected observation poses. We extract
a triangle mesh, map the captured RGB images onto its surface as texture,
and export it in metric object coordinates. These instance-specific assets
provide the known geometry for FoundationPose's six-degree-of-freedom
(6-DoF) tracking in the interaction recordings.
\par
\endgroup

\subsection{Comparison-table definitions and sources}
\label{app:comparison}

Table~\ref{tab:dataset_comparison} compares resources, not pose accuracy.
\ding{51} means documented support; \ding{55} means not satisfied or not
established, not uniformly verified absence. NR is unreported, not zero.
Full-body 3D covers torso, arms, and legs, without requiring fingers or a
specific mesh model. Object pose means dynamic-object trajectories, not
static scene assets. Scene geom. includes layouts, scanned/artist-built assets,
or partial point clouds. Scene mesh requires explicit surface meshes for the
environment, including aligned static-furniture meshes; isolated manipulated-object
meshes alone do not qualify. This distinguishes a provided surface representation
from point-only scene geometry, without imposing a watertightness requirement.
Semantics includes prompts, intents, action/sequence labels, and descriptions of
unequal granularity. Sequence counts retain their source units; multi-view images
are not multiplied into capture hours. Envs. counts physical capture spaces
where established; HOI-M$^3$'s entry of five denotes scene types, not a count
of physical rooms. \DATASET{} has 16 distinct indoor rooms.

\vspace{-1ex}
\paragraph{Acquisition and release scope.}
Uninstr.\ excludes body/object tracking markers, IMUs, and mocap suits/gloves,
but not capture cameras, headset localization sensors, or calibration-only
targets. MMHOI's AprilTags calibrate cameras rather than track the interaction
objects~\citep{kogashi2026mmhoi}. PROX denotes its 60-sequence, 12-room qualitative
release, with scanned scene meshes, not the separate Vicon-referenced quantitative subset
\citep{hassan2019prox}. RICH has five mostly outdoor scanned scene meshes; its semantic
check refers to action-bearing sequence names, not interval captions
\citep{huang2022rich}. Neither provides paired dynamic-object trajectories.

We retain the reported capture modes of GRAB and BEHAVE
\citep{taheri2020grab,bhatnagar2022behave}, InterCap and CHAIRS
\citep{huang2024intercap,jiang2023chairs}, and OMOMO and IMHD$^2$
\citep{li2023omomo,zhao2024imhoi}. InterCap uses the expanded 223-sequence journal
version; CHAIRS uses the final ICCV 17.3-hour release rather than its earlier
preprint. Sparse annotation instants are not converted to unsupported hours.
HOI-M$^3$ uses original CVPR counts, HIMO distinguishes whole-sequence and
segmented language, and ParaHome has a common room setup. HOI-M$^3$ and ParaHome
provide scanned, placed furniture assets; HUMOTO provides artist-built scene
assets. These qualify as scene meshes under the same surface-representation
criterion, without relabeling artist-built assets as measured scans
\citep{zhang2024hoim3,lv2024himo,kim2025parahome,lu2025humoto}.

ACE-Data-0 reports two capture environments and tracked object meshes; an
explicit background-scene surface mesh is not established in its documented
release. HiPHI's 245.7 hours are a mirrored
HOI subset, not unaugmented capture time
\citep{cao2026acedata,ji2026hiphi}. InterAct/-X consolidates seven resources:
11,350/16,201 base/augmented clips and 21.81/30.70 hours are not independent
new acquisitions; mixed instrumentation explains its cross
\citep{xu2025interact}. HOI4D spans 610 rooms with hand-focused supervision and
reconstructs static scene point clouds after masking moving content;
HOT3D's scene clouds refer to Aria recordings in one laboratory. Both receive
Scene geom. checks; their documented point clouds do not establish Scene mesh
support. \DATASET{} supplies reconstructed room meshes aligned with human/object
trajectories across 16 rooms. EgoDex uses
the March 2026 ICLR-associated author version, with upper-body/hand rather
than full-body/object trajectories
\citep{liu2022hoi4d,banerjee2025hot3d,hoque2026egodex}.

\subsection{Language and continuous multimodal example}
\label{app:language}
\label{app:continuous_example}

Normalization lowercases word/number tokens and joins them with single spaces,
preserving internal apostrophes and decimal points but ignoring other punctuation.
There are 12,272 distinct normalized descriptions in the full corpus and
7,856 in the dynamic subset. At a matched sample budget of 1,000 descriptions
without replacement, the expected distinct-string counts are 944.18 for
\DATASET{} and 102.03 for the official OMOMO text archive (4,912 records, 103 unique
strings). 

\begin{figure}[!tb]
\centering
\includegraphics[width=\linewidth, height=0.65\linewidth]{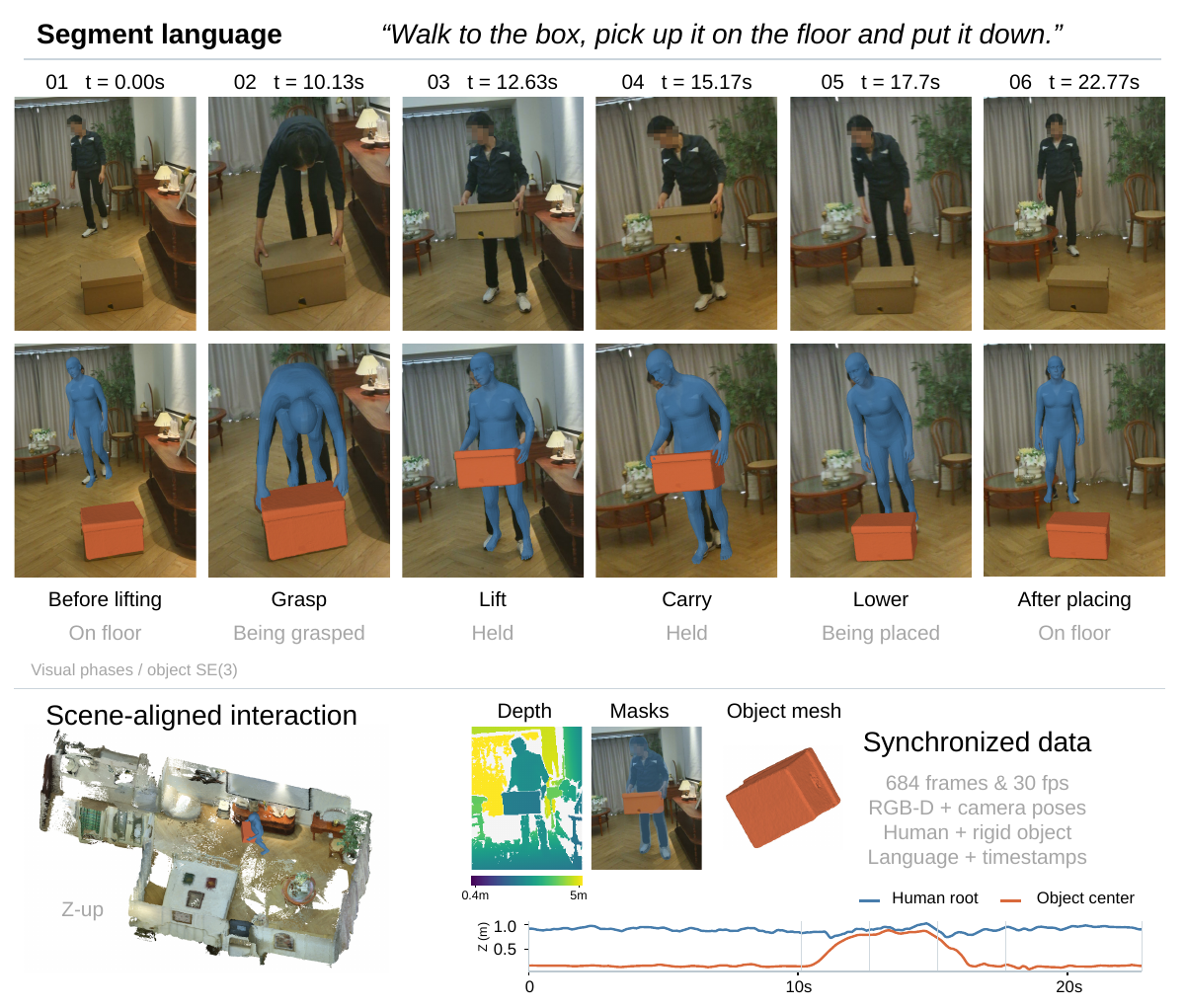}
\caption{One continuous 684-frame interaction, with six real moments spanning
22.77 s. Saved SMPL-X/object estimates, original interval text, scene, depth,
masks, mesh, and height traces share the source timeline. Phase/state labels
are visual interpretations, not extra annotations. Insets use $t=12.63$ s;
depth spans 0.4--5 m, with invalid pixels gray.}
\label{fig:dataset_trajectory}
\end{figure}

Figure~\ref{fig:dataset_trajectory} uses annotation 2, frames 610--1293 at
30 FPS; displayed frames are 610, 914, 989, 1065, 1141, and 1293, with modality
insets at 989 and manual clicks at keyframe 1293. All 684 estimated frames contribute to the traces.

\section{Pipeline implementation details}
\label{app:pipeline}

These are current ordinary-path defaults; archived experiments retain their
recorded configurations and source versions.

\subsection{Scene reconstruction and alignment}
\label{app:scene_details}

For static SfM, HLoc retrieves up to 40 neighbors plus ten-image sequential overlap, using
SuperPoint, SuperGlue, NetVLAD, and COLMAP
\citep{detone2018superpoint,sarlin2020superglue,arandjelovic2016netvlad,
sarlin2019hloc,schonberger2016sfm}. We reject outlying measured-to-SfM depth
ratios using interquartile-range filtering and take the median retained ratio
as the global metric scale.
Fusion defaults use 0.35--7.6 m depth, 0.012 m voxels, and at least three
observations; discontinuity filtering suppresses mixed-depth boundaries.
Camera orientations and reliable IMU readings estimate the vertical direction.
Random sample consensus (RANSAC) fits the floor plane; rotation and translation
align the upward direction to $+z$ and the floor to $z=0$.
For interaction recordings, the default pipeline samples one query every 30
frames. HLoc retrieves up to 30 static images per query and matches
SuperPoint features with SuperGlue to obtain 2D-3D correspondences.
Its PyCOLMAP backend performs RANSAC-based PnP with a 12-pixel reprojection
threshold, followed by pose refinement. Retrieved images are grouped by
shared landmarks, and the pose with the most inliers is retained.
After metric scene alignment and anchor filtering, translation/SLERP
interpolation fills intervening frames; poses outside the anchor span use
the nearest endpoint. Translation smoothing uses a nine-frame window.
Exported frame IDs and calibration are retained.

\subsection{Human fitting}
\label{app:human_details}

A 32D VPoser pose code $z_t=z_t^0+\delta z_t$ parameterizes the body, with
initial code $z_t^0$ and optimized correction $\delta z_t$
\citep{pavlakos2019smplx}. Root translation/orientation and hands receive
corrections; shape, jaw, and eyes stay fixed. The fit maps 63 RTMW body, foot,
and hand landmarks to SMPL-X,
discarding confidence below 0.70 and weighting hand landmarks by 0.65.

\vspace{-1ex}
\paragraph{Observation losses.}
For $\rho_\sigma(r)=\sigma^2r^2/(r^2+\sigma^2)$, image diagonal $d$, projected
landmarks $\hat u_{tj}$, detections $u_{tj}$, and confidence/group/positive-depth
weights $w_{tj}$, the reprojection term is
\begin{equation}
E_{2D}=\frac{\sum_{t,j}w_{tj}\sum_{a\in\{x,y\}}
\rho_{35/d}((\hat u_{tj,a}-u_{tj,a})/d)}
{\max(2\sum_{t,j}w_{tj},1)}.
\label{eq:keypoint_loss}
\end{equation}
The penalty is componentwise, not applied to the 2D residual norm.
Fixed person-mask/depth samples $P_t$ and sampled body vertices
$\widetilde V_t$ give
\begin{equation}
E_D=\operatorname{mean}_{t,p\in P_t}\rho_{.08}\!
\left(\min_{v\in\widetilde V_t}\|p-v\|_2\right)
+.2\operatorname{mean}_{t,v\in\widetilde V_t}\rho_{.08}\!
\left(\min_{p\in P_t}\|v-p\|_2\right).
\label{eq:scan_loss}
\end{equation}
Distances are in meters; the reverse term is weaker because the scan is partial.
Defaults sample supported frames at stride eight, 64 scan points and 128
vertices per frame, depth 0.3--5 m, and initial target distance at most 0.20 m.
This is nearest-vertex agreement, not exact triangle-surface distance.

\vspace{-1ex}
\paragraph{Temporal terms, priors, and schedule.}
Let $b_t,h_t,o_t$ denote body, hand, and root-orientation axis angles,
$\mathcal Y=\{z,b,o,h,q\}$ the preserved parameters, and $y^0$ their
initial values. With $\langle\cdot\rangle$ averaging over frames and scalar
entries, the remaining terms are
\begin{equation}
\begin{aligned}
E_{\mathrm{prior}} &=
 \alpha_z\langle z^2\rangle+\alpha_h\langle h^2\rangle
 +\alpha_a\langle \exp(s_k b_{t,i_k})\rangle_{t,k},\qquad
E_{\mathrm{preserve}} =
 \sum_{y\in\mathcal Y}\mu_y\langle(y-y^0)^2\rangle,\\
E_{\Delta^2 J_t} &=
 \frac{\sum_{t,j}w_j\|\Delta^2J_{t,j}\|_2^2}
 {3(T-2)\sum_jw_j},\qquad
 E_{\Delta^2 x_t}=\langle(\Delta^2x_t)^2\rangle,\quad x\in\{V,R,q,z\}.
\end{aligned}
\label{eq:human_regularizers}
\end{equation}
Here $T$ is the frame count and $w_j$ the joint weights. The elbow/knee
prior uses zero-based body-pose indices $i=(52,55,9,12)$ with signs
$s=(1,-1,-1,-1)$; $\alpha,\mu$ are the prior/preservation weights below.
Joint second differences use selected outputs among the first 76 SMPL-X
joints, with body/hand weights 1/1.35; the vertex term samples 192 vertices.
Translation and latent terms use elementwise second differences; rotation
uses rotation-matrix entry differences, excluding fixed jaw/eye slots.
These losses differ from the body-22 evaluation metric.
Temporal weights for joints, vertices, rotations, translation, and latent
are 180, 20, .25, 50, and .20. Latent/angle/hand prior weights are .005/.005/.01;
initialization-preservation weights for latent, body, root, hands, and
translation are .40, .80, 1, 1, and 1. Image and scan weights are 320 and 30.
Adam runs 70 iterations: 15 at .006 without scan loss, then 55 at .0025 with
scan loss, with gradient norm clipped to eight and bounded corrections.
Removing scan or reprojection loss does not remove depth or image information
from initialization.

\subsection{Object smoothing and tracking safeguards}
\label{app:object_details}
\label{app:tracking_details}

Object updates are $Q_t=\operatorname{Exp}(a_t\delta r_t)Q_t^0$ and
$p_t=p_t^0+a_t\delta p_t$, where $Q_t^0,p_t^0$ are input poses and the binary
update mask $a_t$ is one on optimized frames and zero on inactive/held frames.
Rotations are left-composed in scene coordinates; angular smoothness uses
differences of $\omega_t=\operatorname{Log}(Q_{t+1}Q_t^\top)$ rather than
global axis-angle second differences. Adam uses 100 iterations at .012,
translation/angular temporal weights 80/6, and preservation weights 1/1.
Correction norms are bounded by $s_p=.05$ m/$s_r=.25$ rad; preservation normalizes by
these scales. No human, contact, silhouette, depth, or mesh-fitting term enters
this smoother.

Actual keyframe clicks and known meshes initialize SAM 2 masks and
FoundationPose poses~\citep{ravi2025sam2,wen2024foundationpose}.
We register candidates on visible frames near the annotated keyframe and
select the initialization by mask overlap, depth agreement, and image-boundary
truncation, then track forward and backward. Low mask overlap and reappearance
after occlusion trigger re-registration. Replacement requires improved mask
agreement and satisfactory depth, visibility, reliable-coverage, and
trajectory-continuity checks. Rejected visible poses are interpolated only
between reliable estimates; invisible spans hold the last pose and are marked
unobserved. The smoother leaves these held poses unchanged.

\vspace{-1ex}
\paragraph{Quality-control checks.}
\label{app:quality_control}
Integrity checks align arrays by source-frame ID and verify finite entries,
homogeneous transform rows, rotation orthogonality, and unit determinants.
Camera coverage uses the retained-anchor count, the fractions before the
first and after the last anchor, and the span between them relative to the
segment length. At most one anchor, an uncovered prefix/suffix above 50\%,
or a covered span below 25\% raises a failure flag. Anchor residuals measure
the Euclidean displacement between assembled and localized camera positions
at anchor frames; a 90th-percentile residual above 0.75 m is flagged.
Motion checks measure consecutive translation distances and rotation angles.
Camera/human-root/object translation steps above 0.2/0.25/0.3 m trigger
warnings and above 0.5/2.0/0.8 m trigger failure flags; camera/object rotation
steps use 30/45-degree warning and 60/90-degree failure thresholds.
Geometric diagnostics use minimum mesh height relative to the floor and
nearest-surface distances to the room geometry above the floor; body/object
vertices below $-0.03$ m trigger floor warnings. Object-projection checks
measure rendered-mask overlap, the rendered silhouette fraction inside the
mask, normalized center separation, and depth disagreement; their support
requirements and frame-to-segment decision rule appear in
Appendix~\ref{app:metrics}.

\section{Validation protocols and extended results}
\label{app:validation}

\subsection{Metrics and statistical unit}
\label{app:metrics}

\paragraph{Independent MoCap capture setting.}
To measure absolute 3D reconstruction accuracy, we collect approximately
30 minutes of synchronized RGB-D and MoCap interactions from two participants
(Table~\ref{tab:mocap_accuracy}). Validation takes place in original
\DATASET{} capture environments and reuses the same physical object instances
from dataset collection. This ties the reference measurements directly to
the scene geometry and object shapes and scales encountered during dataset
construction. We process the RGB-D recordings with the standard pipeline;
MoCap trajectories serve as an independent evaluation reference. Fixed
spatial calibration and time synchronization align the two trajectory sets,
using matched body joints and the same object-mesh origins. All configurations
share this calibration and the valid evaluation samples, enabling matched
comparisons of pipeline components under the dataset's capture conditions.
For Table~\ref{tab:mocap_accuracy}, each metric is first averaged over valid
frames within a sequence, then averaged equally across sequences.

G-MPJPE averages Euclidean position errors over the 22 SMPL-X body joints
(indices 0--21, including the root), in the calibrated scene frame.
RA-MPJPE uses the same joints after subtracting each trajectory's own
root position, without rotation or scale alignment. Object translation
error compares the common mesh origin. Object rotation error is the
geodesic angle $\arccos((\operatorname{tr}(\hat R_t^\top R_t^{\rm ref})-1)/2)$
in degrees, with the argument clipped to $[-1,1]$.
For wrist $w\in\{L,R\}$ (joint indices 20/21), define its object-local
position as $r_{t,w}=R_t^\top(J_{t,w}-p_t)$, where $(p_t,R_t)$ is the object
pose. HOI-relative averages $\|\hat r_{t,w}-r_{t,w}^{\rm ref}\|_2$ over both
wrists and valid frames, using the estimated and reference object poses
for their respective coordinate transforms. Position errors are reported in mm.

Table~\ref{tab:pipeline_ablation} uses 390 short clips and 130 long clips
for human fitting, 140 stress clips for tracking, and 900 candidate decisions
for the acceptance guard. The object audit contains 156 clips: 104 exceed
both second-order thresholds below, while 52 fall at or below at least one.
Relative smoothing reductions use the 104-clip group to avoid division by
near-zero initial scores; strict geometric checks cover all 156 clips.
Short human clips are hash-selected within room and low/mid/high initial-jitter
strata; long clips use room-wise hash selection, excluding prior clips.
Tracking stress clips are flagged for weak reliable-update coverage,
poor box overlap, or large pose jumps. Each comparison fixes inputs,
initialization, and evaluation frames.

For 22 body joints (index $j$, coordinate $c$), the temporal score is
$\operatorname{p95}_t\sqrt{\sum_{j,c}(\Delta^2J_{tjc})^2/(22\cdot3)}$.
Objects use $\operatorname{p95}_t\|\Delta^2p_t\|_2$ and
$\operatorname{p95}_t\|\omega_{t+1}-\omega_t\|_2$.
Differences are not divided by physical $\Delta t^2$: units are mm/frame$^2$
or rad/frame$^2$, requiring matched recording rates. Human scan distance takes nearest-vertex distances on fixed depth
points, then frame and clip medians, unlike Equation~\ref{eq:scan_loss}.
Evaluation indices exclude the every-eighth-frame scan-loss samples.
Silhouette and bounding-box IoU remain separate; paired object depth uses
common supported pixels. 

\vspace{-1ex}
\paragraph{Strict geometric rule.}
At silhouette render scale .25, weak overlap requires box IoU $<.08$ and
silhouette IoU $<.06$. Spatial inconsistency means silhouette-inside-mask
fraction $<.30$ or center distance divided by mask-box diagonal $>.75$
(diagonal clamped to one pixel). Depth inconsistency requires at least 40
points and median absolute residual $>.18$ m. Severe separation requires
inside fraction $<.10$ and normalized center distance $>.75$.
A bad frame has severe separation, or spatial inconsistency together with
weak overlap or depth inconsistency. Fewer than two supported frames is
\emph{insufficient}; otherwise at least two bad frames and a bad fraction
$\geq.35$ constitute failure. 

\vspace{-1ex}
\paragraph{Aggregation.}
Continuous metrics in Tables~\ref{tab:pipeline_ablation}
and~\ref{tab:human_full_controls} are cohort medians of clip-level values.
For these metrics, paired benefits in Table~\ref{tab:pipeline_ablation}
are medians of within-clip changes. Relative reduction is
$\operatorname{median}_i[100(c_i-f_i)/c_i]$ for valid positive denominators,
not a ratio of cohort medians. Ordinary-motion clips have initial translation
and rotation temporal p95 above $10^{-6}$ m/frame$^2$ and $10^{-6}$ rad/frame$^2$;
near-zero clips fall at or below either threshold. These measure second-order
variation, not velocity. Repeated configurations do not enlarge the
independent cohort.

\subsection{Component ablations and additional controls}
\label{app:pipeline_ablations}
\label{app:human_controls}

\paragraph{Configuration definitions.}
Table~\ref{tab:pipeline_ablation} compares the full pipeline with matched
ablations or controls using the cohorts and metrics in
Appendix~\ref{app:metrics}. Human temporal ablations remove all five
$E_{\Delta^2x_t}$ terms in Equation~\ref{eq:human_fitting}, or only the joint
term $E_{\Delta^2J_t}$; the long-clip row repeats the all-term ablation.
Depth-scan and keypoint ablations remove $E_D$ and $E_{2D}$, respectively,
while retaining the same initialization. Translation and rotation smoothing
rows measure two outcomes of the same before/after comparison for
Equation~\ref{eq:object_refinement}, rather than separate loss-term removals.
The tracking-safeguard rows evaluate the same control/full comparison in
camera coordinates and after scene assembly; the safeguards are described in
Section~\ref{par:dataset_object_tracking} and Appendix~\ref{app:tracking_details}.
The acceptance-guard row separately replays candidate decisions with and
without replacement screening, counting silhouette-IoU regressions.

\begin{table}[!htb]
\centering
\setlength{\abovecaptionskip}{0pt}
\caption{Effects of human fitting, object smoothing, and tracking safeguards.
Each row compares the full pipeline with a component removed or a control
setting on the same samples. For continuous metrics, Ablated/control and
Full report median clip scores; Paired benefit reports the median improvement
calculated clip by clip. Strict pass counts clips passing geometric checks.
$N$: clips (acceptance guard: decisions); pp: percentage points.}
\label{tab:pipeline_ablation}
\begingroup
\setlength{\tabcolsep}{3pt}
\renewcommand{\arraystretch}{0.90}
\setlength{\aboverulesep}{0.7pt}
\setlength{\belowrulesep}{0.9pt}
\resizebox{\linewidth}{!}{%
\begin{tabular}{llrrrr}
\toprule
Component / setting & $N$ & Metric & Ablated / control & Full & Paired benefit \\
\midrule
Temporal terms & 390 & Body $\Delta^{\text{2}}$ p95 (mm/f\textsuperscript{2}) $\downarrow$ & 91.61 & \textbf{11.88} & 86.41\% $\downarrow$ \\
Joint second order & 390 & Body $\Delta^{\text{2}}$ p95 (mm/f\textsuperscript{2}) $\downarrow$ & 17.78 & \textbf{11.88} & 28.91\% $\downarrow$ \\
Temporal terms (long) & 130 & Body $\Delta^{\text{2}}$ p95 (mm/f\textsuperscript{2}) $\downarrow$ & 85.44 & \textbf{11.08} & 87.03\% $\downarrow$ \\
Depth-scan loss & 390 & Scan distance (mm) $\downarrow$ & 61.82 & \textbf{60.64} & 1.94 mm $\downarrow$ \\
Keypoint loss & 390 & Scan distance (mm) $\downarrow$ & 62.62 & \textbf{60.64} & 2.03 mm $\downarrow$ \\
Keypoint loss & 390 & Person IoU $\uparrow$ & 0.6812 & \textbf{0.6930} & +0.0113 \\
\midrule
Translation smoothing & 104 & Trans. $\Delta^{\text{2}}$ p95 (mm/f\textsuperscript{2}) $\downarrow$ & 19.24 & \textbf{12.39} & 41.44\% $\downarrow$ \\
Rotation smoothing & 104 & Rot. $\Delta\omega$ p95 (rad/f\textsuperscript{2}) $\downarrow$ & 0.196 & \textbf{0.075} & 61.61\% $\downarrow$ \\
\midrule
Tracking safeguards: camera & 140 & Strict pass $\uparrow$ & 82/140 & \textbf{124/140} & +30.0 pp \\
Tracking safeguards: camera & 140 & Object IoU $\uparrow$ & 0.0746 & \textbf{0.2606} & +0.0604 \\
Tracking safeguards: scene & 140 & Strict pass $\uparrow$ & 89/140 & \textbf{128/140} & +27.9 pp \\
Candidate acceptance guard & 900 & IoU regressions $\downarrow$ & 15/900 & \textbf{0/900} & 15 fewer \\
\bottomrule
\end{tabular}%
}
\endgroup
\end{table}

\vspace{-1ex}
\paragraph{Human temporal and observation consistency.}
Temporal regularization reduces body-motion p95 by 86.41\%, with an 87.03\%
reduction on long clips. Depth and keypoint supervision improve agreement
with RGB-D observations (Table~\ref{tab:pipeline_ablation}). These results
complement the independent MoCap comparisons in
Section~\ref{sec:mocap_validation}.

\begin{figure}[!tb]
\centering
\includegraphics[width=\linewidth]{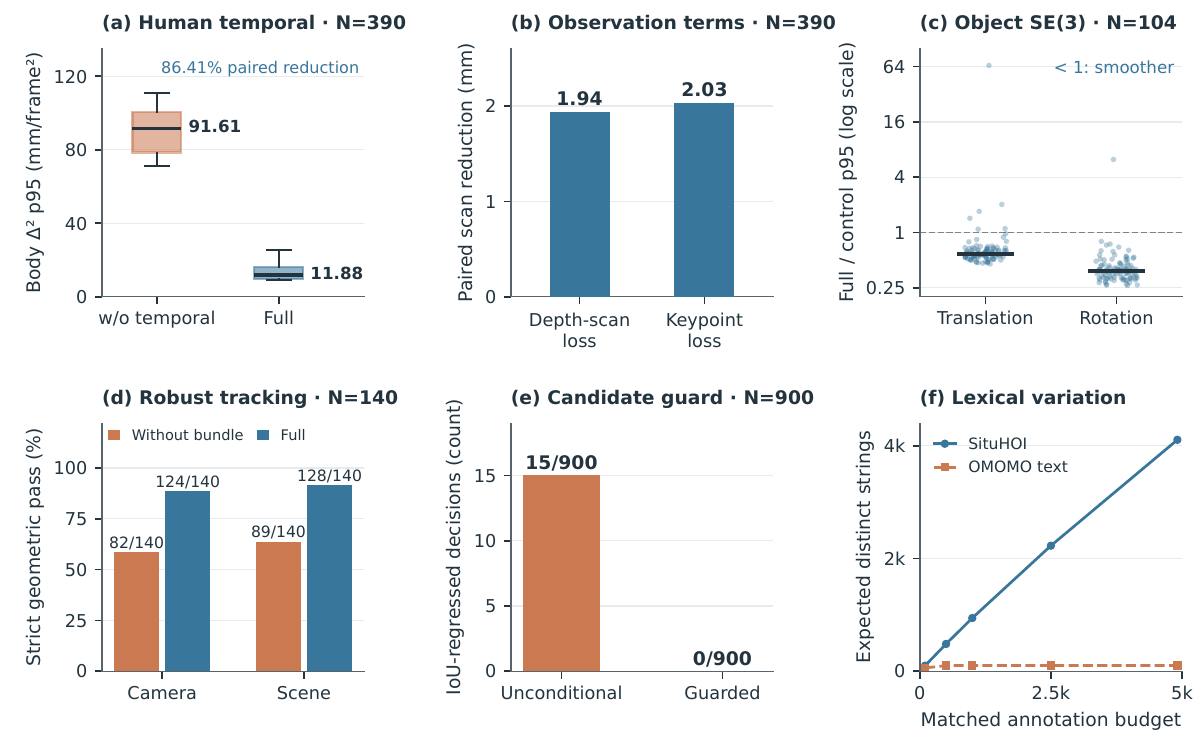}
\caption{Pipeline ablations corresponding to Table~\ref{tab:pipeline_ablation}:
(a) human temporal p95 and (b) median paired scan-distance reductions on 390 clips;
(c) temporal ratios on 104 object clips; (d) strict passes on 140 tracking clips;
(e) IoU regressions over 900 candidate decisions; and (f) matched-budget lexical
variation. Boxes in (a) span the 25th--75th percentiles, with 10th--90th
percentile whiskers; dark bars in (a,c) mark medians.}
\label{fig:pipeline_evidence}
\end{figure}

\vspace{-1ex}
\paragraph{Object refinement and tracking safeguards.}
SE(3) smoothing reduces translation/rotation p95 by 41.44\%/61.61\% on the
104 moving clips. On difficult RGB-D clips, tracking safeguards raise
automated pass counts from 82 to 124 of 140 before scene assembly and from
89 to 128 afterward. The acceptance guard reduces IoU regressions from 15
to zero over 900 archived candidate decisions.
Figure~\ref{fig:pipeline_evidence} summarizes the component comparisons.

\vspace{-1ex}
\paragraph{Human-fitting component and weight ablations.}

Table~\ref{tab:human_full_controls} reports component and temporal-weight
ablations of human fitting, extending Table~\ref{tab:pipeline_ablation}
on the same 390 short clips, with replication on 130 long clips.
All configurations within each cohort use matched evaluation frames;
scan distances are evaluated on held-out depth observations.

\begin{table}[tbp]
\centering
\caption{Human-fitting component and weight ablations on 390 short clips and 130 long clips:
cohort medians of per-clip metrics.
Temporal p95 is in mm/frame\textsuperscript{2}, scan distance in mm, and keypoint residual in
pixels.}
\label{tab:human_full_controls}
\small
\setlength{\tabcolsep}{4pt}
\begin{tabular}{@{}lrrrrr@{}}
\toprule
Configuration & $N$ & Body p95 $\downarrow$ & Scan $\downarrow$ & IoU $\uparrow$ & 2D $\downarrow$ \\
\midrule
\multicolumn{6}{@{}l}{\textit{Fixed short-clip cohort}} \\
Full fitting & 390 & 11.88 & 60.64 & 0.6930 & 20.48 \\
No second-order terms & 390 & 91.61 & 53.47 & 0.6774 & 14.18 \\
No joint second-order term & 390 & 17.78 & 54.25 & 0.6781 & 20.91 \\
Quarter temporal weight & 390 & 19.05 & 53.41 & 0.6644 & 21.41 \\
Half temporal weight & 390 & 14.77 & 61.95 & 0.6804 & 26.11 \\
No scan loss & 390 & 12.29 & 61.82 & 0.6831 & 29.12 \\
No keypoint loss & 390 & 13.78 & 62.62 & 0.6812 & 37.52 \\
\midrule
\multicolumn{6}{@{}l}{\textit{Long-clip replication}} \\
Full fitting & 130 & 11.08 & 46.75 & 0.7175 & 18.05 \\
No second-order terms & 130 & 85.44 & 46.58 & 0.6993 & 17.30 \\
\bottomrule
\end{tabular}
\end{table}

Increasing temporal weight from quarter to half to full reduces body p95
from 19.05 to 14.77 to 11.88 mm/frame$^2$, while person IoU rises from
0.6644 to 0.6804 to 0.6930. Restoring the scan loss reduces median scan
distance from 61.82 to 60.64 mm; restoring keypoint supervision reduces
the median 2D residual from 37.52 to 20.48 pixels. On 130 long clips,
full fitting reduces body p95 from 85.44 to 11.08 mm/frame$^2$ and raises
person IoU from 0.6993 to 0.7175, extending the temporal and silhouette
improvements to longer sequences.

\section{\BENCHMARK\ and \METHOD\ details}
\label{app:benchmark}

\subsection{Temporal queries and reproducible partition}

Eligible annotated segments have paired finite human/object trajectories and
an object mesh. We extract full 135-frame windows at 30 Hz, with strides of
60 frames for TRAIN/HOLDOUT and 135 for VAL, appending a final tail-aligned
window. A window is rejected if either root or object translation deviates
by $\geq5$ m from its first position or moves $\geq1$ m between cached steps.

Each source window is stored as 72 precomputed motion states (the cache),
which are interpolated at the requested query times. The cache uses source times
$\operatorname{rint}(\operatorname{linspace}(0,134,72))/30$, not uniform
16-Hz indices. The observation boundary is a cached anchor; historical
interpolation brackets never cross it. At query rate $\nu=16$ Hz,
$n_H=\max(2,\lceil\nu H\rceil+1)$ points span $[-H,0]$ and
$n_F=\max(1,\lceil\nu F\rceil)$ points span $(0,F]$, including both requested
endpoints. Thus nominal rate does not imply exact $1/16$ s spacing for every
duration. The interface requires $H,F\geq1/30$ s, $H+F\leq4$ s, cache support,
and at most 160 points. The four-second budget leaves anchor-selection margin
within the source windows. Longer supports can be constructed from source
recordings by resampling and retraining; they are not evaluated here.

Training samples legal anchors, then continuous durations, with 20\%
full-span boundary draws where feasible. It excludes
$[.75,.95]\times[1.4,1.8]$ and $[1.4,1.65]\times[.65,.9]$ in $(H,F)$ seconds.
One duration request is shared within a batch; weights are shared across
batches. The first six rows of Table~\ref{tab:benchmark_durations} form the
VAL selection grid; the marked pairs probe withheld duration combinations.
The final three are additional checks. 

The full-corpus protocol uses 12 physical rooms for TRAIN/HOLDOUT and the
remaining four rooms for VAL and checkpoint selection. Within each of the
12 rooms, recordings are deterministically ranked by hash; 20\%, rounded up,
are assigned to HOLDOUT while retaining at least one TRAIN recording.
All windows from a recording stay in one partition. Thus TRAIN and HOLDOUT
share room identities but have disjoint recordings; VAL is room-disjoint
from both. Room counts refer to physical rooms, not separate scans.

\subsection{\METHOD: model and training details}
\label{app:benchmark_implementation}

The 146D physical state contains root translation (3), body rotations
($22\times6$), object translation/rotation (3/6), and wrist distances (2).
The 6D encoding stores the first two rotation-matrix columns; each object's
256 points are fixed-seed samples of its mesh vertices in mesh coordinates.
The shared network predicts 143 residual channels, replacing object 6D
rotation by a three-dimensional rotation-vector correction, right-multiplied
onto the goal-bridge rotation after applying $\operatorname{Exp}$.
Neutral SMPL-X forward kinematics recomputes wrists 20/21 after interpolation.
Five memory tokens encode initial state, goal, durations, object, and scene;
point MLPs pool max/mean geometry features. Attention has six heads,
feed-forward width 768, and dropout .1. With $t,H,F$ measured in seconds,
the common 22D time vector contains $t/4$, phase $t/H$ for history or $t/F$
for future, $H/4$, $F/4$, two history/future indicators, and
$\sin(2\pi ft),\cos(2\pi ft)$ at frequencies
$f\in\{.25,.5,1,2,4,8,12,16\}$ Hz.

Reconstruction weights for root, body, object position, object rotation,
and distances are 4, 1, 8, 1, and .2. Reconstruction uses mean squared error
on normalized residual channels. Velocity error uses consecutive differences
divided by query-time spacing, scaled to a $1/16$ s reference step, with weight .2.
AdamW uses batch 128, learning rate $3\cdot10^{-4}$, weight decay $10^{-4}$,
cosine decay, gradient clipping at one, and 120 epochs. Three seeds are
20260910/11/12. Checkpoints minimize mean normalized VAL reconstruction over
the six selection pairs, checked every ten epochs.

For the object-translation refinement head, let $p_0,R_0$ be the last observed object
pose, $p_F,R_F$ the goal, $v_0,\omega_0$ the observed linear and angular
velocities, $u=t/F$, $\delta=\operatorname{Log}(R_0^\top R_F)$, and
$b(t)=\frac{\tau}{\tau+F}Fu(1-u)^2$, with $\tau=1$ s. Then
\begin{align}
p_{\rm prior}(t)&=(1-u)p_0+up_F+b(t)\bigl(v_0-(p_F-p_0)/F\bigr),\nonumber\\
\hat R(t)&=R_0\operatorname{Exp}\!\left[u\delta+
b(t)(\omega_0-\delta/F)\right].
\label{eq:benchmark_priors}
\end{align}
Least-squares velocity uses the last .2 s (all history if fewer than two
points), clipped to 2 m/s and $2\pi$ rad/s. The 225--128--128--3 MLP uses
sigmoid-weighted linear unit (SiLU) activations and
combines frozen 192D features, 22D time features, goal displacement, velocity,
anchor position, and two observed wrist distances; its last layer starts at
zero. Its dimensionless output $r_\phi(t)$ refines translation as
\begin{equation}
\hat p(t)=p_{\rm prior}(t)+4u(1-u)c(F)r_\phi(t),
\qquad c(F)=0.1\min\!\left(\frac{F}{1\,\mathrm{s}},1\right)\mathrm{m}.
\label{eq:benchmark_refinement}
\end{equation}
The envelope vanishes at $t=0,F$, preserving both endpoints; $c(F)$ reduces
the correction scale for short forecasts. Rotation follows
Equation~\ref{eq:benchmark_priors}. Decoding restores scene coordinates,
observed history, and the supplied object endpoint.
Its loss is intermediate object ADE plus .01 times squared metric
translation residual. The shared predictor stays frozen during 40 AdamW
epochs (learning rate $3\cdot10^{-4}$, weight decay .01).
Head selection checks six-profile VAL ADE every five epochs with
proximity/rotation guards,
selecting epochs 40/30/40. These extra epochs are not a compute-matched
advantage claim.

\vspace{-1ex}
\paragraph{Baseline implementations.}
The compared methods were developed for different conditioning and prediction
settings, and their original interfaces do not directly match \BENCHMARK{}.
We therefore adapt the authors' core implementations of CHOIS, DecHOI,
InterDiff, and OMOMO to a common scene-grounded, goal-conditioned joint
prediction interface with variable history and forecast durations, while
retaining their core temporal architectures.
A shared motion representation and geometry/goal interface align the available
information and evaluation targets. The same resampled query times and 22D physical-time features
explicitly encode elapsed seconds and the requested history/forecast durations,
so each model can distinguish different temporal requests within one set of
weights. For OMOMO, a bridge from the last observed object pose to the supplied
goal provides the trajectory conditioning required by its hand-to-body stages
using only the benchmark inputs. The adapted cores retain their respective
reconstruction/denoising objectives on the common trajectory targets. To focus
the comparison on this shared prediction task, method-specific auxiliary
losses and contact/physics refinement remain outside its scope. These
adaptations support consistent evaluation across model architectures, input
durations, and forecast horizons.

\subsection{Evaluation metrics and duration-wise results}
\label{app:benchmark_body_metrics}

Root and object average displacement errors (ADE) measure Euclidean
translation error against the fitted dataset trajectories, in mm.
Object translation and geodesic rotation errors exclude the supplied
endpoint $t=F$; root, body-rotation, and proximity metrics include it.
Errors are averaged over valid future queries within each window, then
equally across windows.

For each wrist and query, proximity is positive when its minimum distance
to the 256 object samples is below 0.20 m. Predicted and reference states
are computed from their respective human/object geometry, rather than the
model's auxiliary distance channels. Within each duration pair and seed,
true positives (TP), false positives (FP), and false negatives (FN) are
pooled over both wrists, future queries, and all evaluated windows;
$\mathrm{F1}=100\cdot2\mathrm{TP}/(2\mathrm{TP}+\mathrm{FP}+\mathrm{FN})$.
An empty denominator is undefined. The ``All 11 pairs'' columns in
Table~\ref{tab:benchmark_main} equally average pair-level scores within
each seed, then report the mean $\pm$ sample standard deviation across
the three training seeds. Individual duration columns use the same
seed-level summary.

The deterministic Hold reference retains the last observed human root
and body pose, while linearly interpolating object translation and using
SLERP for object rotation toward the supplied goal.
Under the same eleven-pair aggregation as Table~\ref{tab:benchmark_main},
rotation errors (degrees) are 12.51 for Hold, $12.53\pm0.00$ for CHOIS,
$12.53\pm0.00$ for DecHOI, $14.26\pm0.14$ for InterDiff,
$12.51\pm0.00$ for OMOMO, and $12.47\pm0.00$ for \METHOD.
Body-rotation error averages geodesic angles over all future queries and
22 joints: the root uses scene-frame orientation and the remaining joints
use parent-relative rotations. Means over seeds and the eleven pairs are
7.36/6.84/6.94/6.90/7.88/5.68 degrees for Hold/CHOIS/DecHOI/InterDiff/OMOMO/\METHOD. Across the six selection pairs on VAL, object ADE is 55.736 mm for the
unrefined predictor, 49.529 for a goal bridge, 47.887 after velocity
correction, 47.818 with a kinematics-only head, and 47.466 with frozen shared features.

\begin{table}[!htb]
\centering
\caption{\METHOD\ across all 11 duration pairs on all 6,948 HOLDOUT windows.
Entries are three-seed means; object ADE also shows sample SD. Daggers mark
pairs inside excluded training-duration regions.}
\label{tab:benchmark_durations}
\begingroup
\setlength{\tabcolsep}{5pt}
\begin{tabular}{@{}rrrrrr@{}}
\toprule
$H$ (s) & $F$ (s) & Root (mm) & Object (mm) & Rotation (deg) & F1 (\%) \\
\midrule
0.25 & 0.50 & 102.09 & 21.43\textpm0.01 & 5.54 & 85.75 \\
0.50 & 1.00 & 145.45 & 45.13\textpm0.06 & 9.39 & 79.55 \\
0.50 & 3.50 & 331.21 & 123.83\textpm0.70 & 23.36 & 58.90 \\
1.00 & 2.00 & 217.63 & 76.68\textpm0.26 & 15.93 & 69.29 \\
2.00 & 1.00 & 153.42 & 43.07\textpm0.05 & 9.19 & 78.25 \\
3.00 & 0.50 & 96.28 & 19.68\textpm0.02 & 5.23 & 85.30 \\
\midrule
0.85\textsuperscript{\textdagger} & 1.60 & 186.86 & 64.83\textpm0.22 & 13.37 & 73.14 \\
1.50\textsuperscript{\textdagger} & 0.80 & 131.88 & 33.41\textpm0.05 & 7.64 & 81.03 \\
\midrule
0.25 & 3.75 & 355.04 & 132.92\textpm0.66 & 24.44 & 58.08 \\
3.75 & 0.25 & 64.87 & 9.39\textpm0.00 & 2.77 & 90.93 \\
0.37 & 2.81 & 274.27 & 100.41\textpm0.53 & 20.29 & 63.51 \\
\bottomrule
\end{tabular}
\endgroup
\end{table}

\label{app:benchmark:end}

\section{External-model evaluation details}
\label{app:external_validation}

\vspace{-1ex}
\paragraph{Transfer protocol and adaptations.}
The source subset contains 512 TRAIN segments, selected round-robin from
hash-ranked recording-object groups. Hash selection retains at most 16
non-overlapping 35-frame windows per segment within approved TRAIN ranges,
yielding 6,562 windows. BEHAVE TRAIN/VAL/TEST contain 1,450/2,239/3,602 windows
from 37/50/85 disjoint recordings; TEST uses Date03.
Both domains use a neutral 22-joint skeleton and coordinates anchored to the
first observed root pose. Object meshes are centered by their vertex mean,
with translations adjusted to preserve world geometry; ADE follows this
center's trajectory. InterDiff uses the author SMPL diffusion branch
without interaction correction and 50-step denoising diffusion implicit
model (DDIM) sampling. HOIMotion's discrete cosine transform and graph
convolution networks consume past joints, fitted head direction, and object
box corners, and predict body joints. We evaluate fixed final checkpoints,
averaging errors over future frames and test windows within each seed.
The fixed 3k source-only HOIMotion checkpoints, evaluated with zero BEHAVE
updates, achieve $93.27\pm1.61$ mm MPJPE. Extrapolating the mean of the last
two observed joint displacements gives 123.73 mm; a static-body reference
gives 127.67 mm.

\vspace{-1ex}
\paragraph{Transfer gains and reference conditions.}
Using the reported seed means in Table~\ref{tab:external_transfer},
InterDiff's object ADE decreases from 157.39 to 141.32 mm at the short
budget and from 155.24 to 139.36 mm at the long budget, reductions of
16.07 and 15.88 mm. HOIMotion's body MPJPE decreases from 67.93 to
56.59 mm and from 67.73 to 55.43 mm, reductions of 11.34 and 12.30 mm.
The table reports sample SD across eight seeds for InterDiff and three
for HOIMotion.
Both methods use matched total budgets of 6k and 12k updates. Pretraining
or BEHAVE warm-up lasts 3k updates, followed by 3k or 9k updates on BEHAVE,
respectively. Conditions share initial weights and target batches, with
optimizers reset before target training.

At the long budget, InterDiff's static-object reference is 138.32 mm and
its full-target-training reference (6,741 windows) is 131.16 mm.
These references distinguish the pretraining benefit from
absolute forecasting performance.

\subsection{Language and scene-context ablations}
\label{app:external_context}

TMR~\citep{petrovich2023tmr} evaluates text-motion retrieval and
SceneDiffuser~\citep{huang2023scenediffuser} scene-conditioned prediction.
Removed retains only object-name text or removes background scene points;
Shuffled permutes the corresponding training context within object categories.

\begin{table}[!htb]
\centering
\caption{Language and scene-context ablations on \DATASET{} HOLDOUT.
Values are means $\pm$ sample SD over three runs; bold denotes the best
mean within each row.}
\label{tab:external_context}
\begingroup
\small
\setlength{\tabcolsep}{3pt}
\renewcommand{\arraystretch}{1.0}
\setlength{\aboverulesep}{1pt}
\setlength{\belowrulesep}{1pt}
\begin{tabular*}{\linewidth}{@{\extracolsep{\fill}}lllr>{\columncolor{d2hoirow}}rrr@{}}
\toprule
Method & Context & Metric & Updates & Full & Removed & Shuffled \\
\midrule
TMR & Text & R@1 (\%) $\uparrow$ & 6k & \textbf{8.91\textpm1.21} & 3.11\textpm0.47 & 3.42\textpm0.94 \\
SceneDiffuser & Scene & MPJPE (mm) $\downarrow$ & 12k & \textbf{792.40\textpm11.15} & 803.74\textpm25.11 & 804.93\textpm20.24 \\
& Scene & MPJPE (mm) $\downarrow$ & 50k & \textbf{760.11\textpm18.05} & 809.53\textpm31.61 & 802.52\textpm22.29 \\
\bottomrule
\end{tabular*}
\endgroup
\end{table}

\vspace{-1ex}
\paragraph{Semantic retrieval.}
TMR uses one entry per fitted segment under the recording-level partition,
encoding 64 frames across the complete segment with 22 body joints and eight
object corners. A common 20 m range screen relative to the initial
root excludes one TRAIN and three VAL segments, retaining
3,432/1,365/954 TRAIN/VAL/HOLDOUT segments. All conditions train from scratch
for 6k updates with three seeds. Positives share the normalized full
description and object category; R@1 measures the probability of a positive
top-ranked motion, averaging tied ranks. Shuffling affects training pairs
only. Full descriptions improve HOLDOUT R@1 over object-name and shuffled-text
controls by 5.80/5.49 percentage points (Table~\ref{tab:external_context}),
demonstrating information beyond object identity. Joint seed/recording intervals are
$[2.16,9.64]$/$[1.93,9.19]$, and seed-$t$ intervals are
$[1.63,9.97]$/$[1.72,9.25]$. VAL R@1 is
3.47/1.42/1.76\% in the same condition order, with gain intervals crossing
zero. Uniform same-object retrieval reaches 2.47\% on HOLDOUT.

\vspace{-1ex}
\paragraph{Scene-conditioned forecasting.}
SceneDiffuser uses PointNet to encode background/object points and predicts
body joints and object corners from motion history. Under the recording-level
split, HOLDOUT contains 2,360 forecasting windows, taking at most three
non-overlapping four-second windows per segment.
Full scene context yields the lowest mean MPJPE at both budgets, reducing
it by 6.10\%/5.29\% over Removed/Shuffled at 50k updates
(Table~\ref{tab:external_context}). These separate ablations support the value
of aligned language and geometry.

\label{app:external_validation:end}

\end{document}